\pdfoutput=1

\documentclass[11pt]{article}

\usepackage[]{acl}

\usepackage{times}
\usepackage{latexsym}
\usepackage{times}
\usepackage{latexsym}
\usepackage{multirow}
\usepackage{url}
\usepackage{tabularx}
\usepackage{latexsym}
\usepackage{balance}
\usepackage{bm}
\usepackage{amssymb}
\usepackage{amsmath}
\usepackage{url}
\usepackage{makecell}
\usepackage{multirow}
\usepackage{subfig}
\usepackage{graphicx} 
\usepackage{color}
\usepackage{colortbl}
\usepackage{textcomp}
\usepackage{CJK}
\usepackage{enumitem}
\usepackage{booktabs}
\usepackage{microtype}
\usepackage{arydshln}
\usepackage{amsfonts}
\usepackage{wasysym}
\usepackage{eufrak}
\usepackage{pifont}
\usepackage[normalem]{ulem}
\useunder{\uline}{\ul}{}
\usepackage[T1]{fontenc}

\usepackage[utf8]{inputenc}

\usepackage{microtype}

\title{Revisiting Knowledge Distillation for Autoregressive Language Models}

\author{%
  Qihuang~Zhong$^{1}$,
  Liang~Ding$^{2}$,
  Li Shen$^{3}$,
  \textbf{Juhua~Liu}$^{1}$\thanks{~~Corresponding Authors: Juhua Liu (e-mail: liujuhua@whu.edu.cn), Bo Du (e-mail: dubo@whu.edu.cn)},
  \textbf{Bo~Du}$^{1*}$,
  \textbf{Dacheng~Tao}$^{4}$ \\
  \fontsize{9.0pt}{\baselineskip}\selectfont $^{1}$ School of Computer Science, National Engineering Research Center for Multimedia Software, Institute of Artificial Intelligence \\ 
  \fontsize{9.0pt}{\baselineskip}\selectfont  and Hubei Key Laboratory of Multimedia and Network Communication Engineering, Wuhan University, China \\
  \fontsize{9.0pt}{\baselineskip}\selectfont $^{2}$ The University of Sydney, Australia \quad $^{3}$ Sun Yat-sen University, China \quad $^{4}$ Nanyang Technological University, Singapore \\
\fontsize{9.0pt}{\baselineskip}\selectfont \texttt{\{zhongqihuang, liujuhua, dubo\}@whu.edu.cn}\\
\fontsize{9.0pt}{\baselineskip}\selectfont \texttt{\{mathshenli, liangding.liam, dacheng.tao\}@gmail.com}
}

\begin{document}
\maketitle

\begin{abstract}
Knowledge distillation (KD) is a common approach to compress a teacher model to reduce its inference cost and memory footprint, by training a smaller student model. However, in the context of autoregressive language models (LMs), we empirically find that larger teachers might dramatically result in a poorer student. In response to this problem, we conduct a series of analyses and reveal that \textit{different tokens have different teaching modes}, neglecting which will lead to performance degradation. Motivated by this, we propose a simple yet effective \textbf{adaptive teaching} approach (ATKD) to improve the KD. The core of ATKD is to reduce rote learning and make teaching more diverse and flexible. Extensive experiments on 8 LM tasks show that, with the help of ATKD, various baseline KD methods can achieve consistent and significant performance gains (up to +3.04\% average score) across all model types and sizes. More encouragingly, ATKD can improve the student model generalization effectively.
\end{abstract}
\section{Introduction}
\label{sec_intro}

Autoregressive language models (LMs), such as GPT-4~\cite{openai2023gpt4}, PaLM~\cite{chowdhery2023palm} and LLaMA2~\cite{touvron2023llamav2}, have achieved great success in a numerous tasks~\cite{zhong2023chat,Peng2023ChatGPT4MT,Lu2023EAPrompt}. However, with the scaling of model size, the inference and deployment of these LMs become more computationally expensive and memory intensive, hindering the development of industrial applications. Hence, it is crucial and green to compress these LMs and accelerate the inference, while not losing much performance~\cite{schwartz2020green}.

\begin{figure}[t]
\centering
    \includegraphics[width=0.45\textwidth]{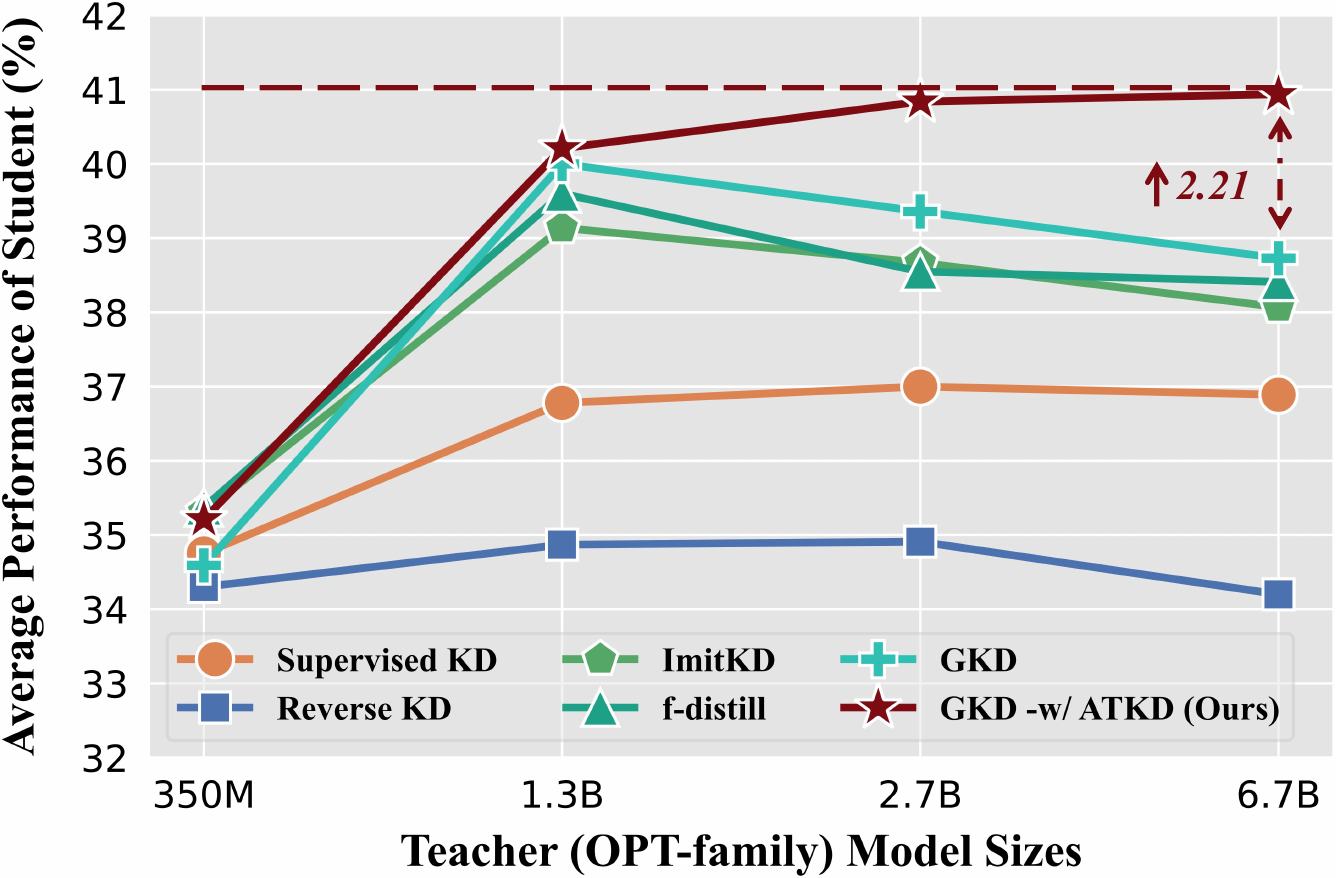}
    \caption{\textbf{Comparisons of different KD methods} for distilling the student (OPT-125M). The x-axis denotes the OPT-based teacher sizes, while the y-axis denotes the average performance of students on $\mathcal{S}_{\text{NLG}}$ and $\mathcal{S}_{\text{NLU}}$. The evaluation details are in \S\ref{sec:experiments}. Notably, ATKD can be combined with various KD methods, and we only report the results of ``GKD + ATKD'' for ease of illustration.}
    \label{fig_1}
\end{figure}

To achieve this goal, a common approach is knowledge distillation (KD), which aims to compress a large teacher model by distilling its knowledge into a small student model~\cite{hinton2015distilling, kim2016sequence}. Recently, in the context of autoregressive LMs, various novel learning algorithms have been proposed to achieve better distillation performance~\cite{wen2023f,agarwal2023gkd}. Despite their remarkable performance, we empirically find a \textbf{counter-intuitive} phenomenon, where \textit{larger teachers might dramatically result in a poorer student, especially when the model capability gap is large}. As illustrated in Figure~\ref{fig_1}, the performance of student degrades when the teachers are too large, which is similar to the findings of~\citet{mirzadeh2020improved, cho2019efficacy,zhang2023lifting}. 

Although a few works aim to investigate this problem and propose to fill the gap, they are mostly studied for vision models~\cite{mirzadeh2020improved, cho2019efficacy} or discriminative language understanding models~\cite{zhang2023lifting}, while the autoregressive KD for generative LMs is yet to be explored. In this work, we investigate this problem from the perspective of the distillation objective, which is at the core of autoregressive KD. Specifically, taking the classical token-level KD objective, \textit{i.e., forward KL-Divergence}, as an example, we first reformulate it as two parts: 1) \textbf{target-oriented knowledge distillation} (\textsc{TKD}), which enforces the student model to learn the target-related information
; 2) \textbf{diversity-oriented knowledge distillation} (\textsc{DKD}), which encourages the student to learn more diverse knowledge from the teacher in the non-target classes. These two parts are tied by a token-wise factor, which reflects the teacher's uncertainty and we denote it as \textbf{uncertainty coefficient} (\textsc{UnC}). After reformulating the distillation objective, we conduct a series of preliminary analyses on the popular OPT-family~\cite{zhang2022opt} models, and find that: 
\begin{itemize}
    \item[]\ding{182} \textsc{UnC} measures the learning difficulties of tokens, where the hard-to-learn ones are more important for KD.
    \item[]\ding{183} \textsc{DKD} contributes more but is greatly suppressed, especially for the larger teachers.
    \item[]\ding{184} \textsc{TKD} plays different roles in tokens with different learning difficulties.
\end{itemize}

Based on these observations, we can conclude that \textbf{different tokens have different teaching modes}, and (one of) the limitations of KD comes from the neglect of this principle. To address this limitation, we propose a simple yet effective adaptive teaching method (referred to as \textbf{ATKD}) to improve the KD. The core of ATKD is to reduce rote learning and make teaching more diverse and flexible. Specifically, ATKD skips the target-oriented teaching for the (less-informative) easy-to-learn tokens and pays more attention to the diverse learning of hard-to-learn tokens.

We evaluate ATKD on a variety of LM benchmarks, including 5 language generation tasks and 3 language understanding tasks, upon 3 types of autoregressive LMs: OPT~\cite{zhang2022opt}, Pythia~\cite{biderman2023pythia}	 and LLaMA~\cite{touvron2023llamav2}. Results show that ATKD can not only alleviate the problem of performance degradation in larger teachers, but also bring consistent and significant improvements (up to +3.04\% average score) into various baseline KD methods among all model types and sizes. Moreover, compared to the standard KD, ATKD can effectively improve the generalization of distilled students.

\paragraph{Contributions.} To summarize, our contributions are three-fold: (1) Our study reveals that \textit{different tokens have different teaching modes}, neglecting which will cause the sub-optimal distillation performance, especially in larger teachers. (2) We propose a simple yet effective, plug-and-play approach (ATKD) to alleviate this problem and improve the quality of teaching. (3) Extensive experiments show that ATKD outperforms the standard KD with up to +3.04\% average gains and improves the student's model generalization effectively.

\section{Rethinking Knowledge Distillation for Autoregressive LMs}
\label{sec:preliminaries}
In this section, we first delve into the mechanism of classic knowledge distillation and then present the empirical analyses of this strategy in detail.

\subsection{Recap of Knowledge Distillation}
\paragraph{Notations.}
For autoregressive LMs, the classic KD aims to approximately minimize Kullback-Leibler (KL) divergence between the teacher and student output distribution at each token~\cite{hinton2015distilling}. Let $\mathbf y = \{\mathrm y_1,...,\mathrm y_T$\} denote the target sequence and $V$ denote the vocabulary, we refer to $\mathbf{y}_{<t}$ as $\{\mathrm y_1,...,\mathrm y_{t-1}$\}, where $t \in \{1,..., T\}$ and $\mathrm y_t \in V$. Specifically, the loss function can be formulated as:
\begin{align*}\label{eq:classicKD}
    \mathcal{L}_{\text{KL}}(\mathbf{p}||\mathbf{q}) &=- \sum_{t=1}^{T}{\text{KL}}\left(\mathbf{p}(\mathrm y_t|\mathbf{y}_{<t})||\mathbf{q}(\mathrm y_t|\mathbf{y}_{<t})\right)\\
    &=- \sum_{t=1}^{T} \mathbf{p}(\mathrm y_t|\mathbf{y}_{<t}) \log \left( \frac{\mathbf{p}(\mathrm y_t|\mathbf{y}_{<t})}{\mathbf{q}(\mathrm y_t|\mathbf{y}_{<t})} \right),
\end{align*}
where $\mathbf{p}=[p_1,...,p_C]$ and $\mathbf{q}=[q_1,...,q_C]$\footnote{For simplicity, we only consider the formulation of $\mathbf{p}$ in the following context. Note that the $\mathbf{q}$ is similar to $\mathbf{p}$.} are the predicted distributions of the teacher and student, respectively; 
$p_{i}$ is the probability of the $i$-th class and $C$ is the number of vocabulary $V$, $\text{KL}$ refers to the KL divergence. For simplicity, we denote $\mathbf{p}(\mathrm y_t|\mathbf{y}_{<t})$ as $\mathbf{p}^t$, and $p^t_i$ as the probability of the $i$-th class at $t$-th step.
Here, $p^t_{i}$ is determined using a softmax function:
\begin{equation}
    p^t_{i} = \frac{\exp(z^t_{i})}{\sum_{j=1}^{C} \exp(z^t_{j})},
\label{defpi}
\end{equation}
where $z^t_i$ represents the logit of the $i$-th class in $V$. Let $g_t$ denote the target token/class at $t$-th step, we can obtain the binary probabilities $\mathbf{p}^t_{\mathbf b}=[p^t_{g_t},p^t_{\backslash g_t}]$, where probability of the target class $p^t_{g_t}$ and non-target classes $p^t_{\backslash g_t}$ can be calculated as:
\begin{equation*}
	p^t_{g_t} = \frac{\exp(z^t_{g_t})}{\sum_{j=1}^{C} \exp(z^t_{j})}, p^t_{\backslash g_t} = \frac{\sum_{k=1,k\neq g_t}^{C} \exp(z^t_{k})}{\sum_{j=1}^{C} \exp(z^t_{j})}.
\end{equation*}
Moreover, for independently analyzing the probabilities among non-target classes, we declare $\mathbf{\hat{p}}^t=[\hat{p}^t_1,...,\hat{p}^t_{g_t-1}, \hat{p}^t_{g_t+1},...,\hat{p}^t_C]$, where $\hat{p}^t_{i}$ is:
\begin{equation}
    \hat{p}^t_{i} = \frac{\exp(z^t_{i})}{\sum_{j=1, j\neq g_t}^{C} \exp(z^t_{j})}.
\label{defpnt}
\end{equation}

\paragraph{Reformulation of $\mathcal{L}_{\text{KL}}$.}
Here, we are inspired by~\citet{zhao2022decoupled}\footnote{Although the reformulation of $\mathcal{L}_{\text{KL}}$ is inspired by the previous work~\cite{zhao2022decoupled}, we take a further step by exploring the potential mechanism of autoregressive KD from the perspective of teaching modes among different tokens, which are our main contributions.}, and attempt to reformulate $\mathcal{L}_{\text{KL}}$ with the binary probabilities $\mathbf{p}^t_{\mathbf b}$ and the probabilities among non-target classes $\hat{\mathbf{p}}^t$, which can be reformulated as:
\vspace{-1pt}
\begin{equation}
    \begin{split}
    \mathcal{L}_{\text{KL}} =- \sum_{t=1}^{T} ( p^t_{g_t}\log(\frac{p^t_{g_t}}{q^t_{g_t}}) + \sum_{j=1,j\neq g_t}^{C} p^t_{j}\log(\frac{p^t_{j}}{q^t_{j}})).
    \end{split}
    \label{kd_ori_form}
\end{equation}
According to Eq.~\ref{defpi} and~\ref{defpnt}, we have $p^t_{i}=\hat{p}^t_{i}*p^t_{\backslash g_t}$, and can further rewrite Eq.~\ref{kd_ori_form} as:
\begin{equation}
    \begin{split}
    \mathcal{L}_{\text{KL}} &=- \sum_{t=1}^{T} \left( p^t_{g_t}\log(\frac{p^t_{g_t}}{q^t_{g_t}}) \right. \\
    &\left. + p^t_{\backslash g_t} \sum_{j=1,j\neq g_t}^{C} \hat{p}^t_{i}\left(\log(\frac{\hat{p}^t_{j}}{\hat{q}^t_{j}})+\log(\frac{p^t_{\backslash g_t}}{q^t_{\backslash g_t}})\right) \right) \\
    &=- \sum_{t=1}^{T} \left(p^t_{g_t}\log(\frac{p^t_{g_t}}{q^t_{g_t}})+p^t_{\backslash g_t}\log(\frac{p^t_{\backslash g_t}}{q^t_{\backslash g_t}})\right.\\
    &\left.+p^t_{\backslash g_t} \sum_{j=1,j\neq g_t}^{C}\hat{p}^t_{i} \log(\frac{\hat{p}^t_{j}}{\hat{q}^t_{j}}) \right) \\
    &=- \sum_{t=1}^{T} \left({\text{KL}}(\mathbf{p}^t_{\mathbf b}||\mathbf{q}^t_{\mathbf b})+p^t_{\backslash g_t}{\text{KL}}(\mathbf{\hat{p}^t}||\mathbf{\hat{q}^t}) \right).
    \end{split}
    \label{kd_refor}
\end{equation}

As seen, we can reformulate the classic KD objective as a combination of binary classification loss on the target class, and KL loss on the non-target classes. The former forces the student to learn the target-related information, and we thus denote it as \textbf{target-oriented knowledge distillation} (\textsc{TKD}). Conversely, the latter encourages the student to distill the diverse knowledge among non-target classes, and we denote it as \textbf{diversity-oriented knowledge distillation} (\textsc{DKD}). Moreover, we find that \textsc{TKD} and \textsc{DKD} are tied by a token-wise factor $p^t_{\backslash g_t}$, which could reflect the teacher's uncertainty on the tokens, \textit{i.e.}, the larger $p^t_{\backslash g_t}$ denotes the more uncertainty\footnote{For example, the token with $p_{\backslash g_t}=0.7$ is more uncertain than the one with $p_{\backslash g_t}=0.1$.} in the teacher output distribution. Hence, we refer to $p^t_{\backslash g_t}$ as \textbf{uncertainty coefficient} (\textsc{UnC}).

\subsection{Empirical Analyses}
 \label{sec:analysis}

\paragraph{Setting.} We conduct experiments by first fine-tuning larger LMs on the instruction-response dataset $\mathcal{D}$ as teachers. Then, we use different KD methods to distill a smaller student on $\mathcal{D}$ with the teacher’s guidance. Here, we use the original OPT-125M as the student and use the other OPT-family models (\textit{i.e.}, OPT-350M/-1.3B/-2.7B/-6.7B) as teachers. Alpaca-GPT4~\cite{peng2023instruction} is used as training data, and the models are evaluated on three instruction-following datasets, \textit{i.e.}, DollyEval~\cite{gu2023knowledge}, VicunaEval~\cite{chiang2023vicuna} and SelfInst~\cite{wang2022self}. We follow~\cite{gu2023knowledge} and use the LLM-based metric, \textit{i.e.}, \textbf{LLM-as-a-Judge}, to quantify the model responses. Specifically, we ask GPT-3.5-Turbo-1106\footnote{The analysis of this evaluator is shown in Appendix~\ref{appendix_chatgpt}.} to compare model responses with the ground-truth answers and raise 1-10 scores for both responses and report the ratio of the total score of model responses and ground-truth answers. 

\begin{figure}[t]
\centering
    \includegraphics[width=0.45\textwidth]{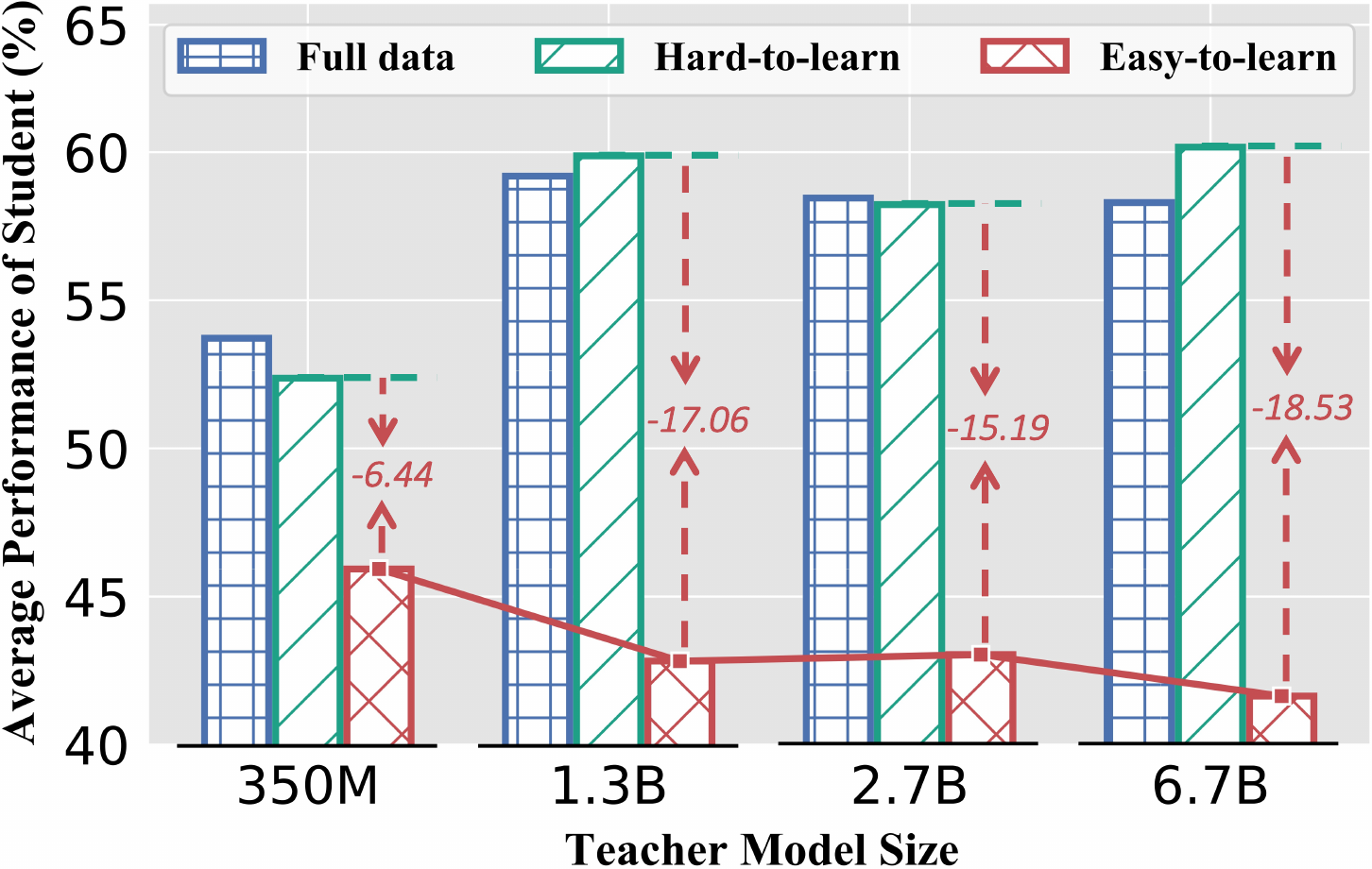}
    \caption{\textbf{Comparisons of different training tokens.} The y-axis denotes the average performance of students (OPT-125M) on the evaluated tasks, while the x-axis denotes the sizes of OPT-based teachers.}
    \label{fig:analysis_1}
\end{figure}

\paragraph{Findings.} To reveal the drawbacks of $\mathcal{L}_{\text{KL}}$ and explore the reasons for performance degradation in large teachers, we conduct systematic analyses to investigate the different effects of \textsc{UnC}, \textsc{TKD} and \textsc{DKD}, respectively. Through the extensive analyses, we empirically observe that: 

\paragraph{\ding{182} \textsc{UnC} measures the learning difficulties of tokens, where the hard-to-learn ones are more important for KD.} 
Motivated by the token imbalance nature and the truth that different tokens in a sequence contribute differently to the sentence meaning~\cite{church1990word,chen2020content}, we conjecture that different tokens play different roles in autoregressive KD. Intuitively, the tokens with less uncertainty have simple learning patterns and \textit{easy-to-learn}, while the more uncertain tokens are more informative and are \textit{hard-to-learn}. To verify our conjecture, we rank the training tokens according to the \textsc{UnC} for each mini-batch and evenly split them into two subsets. For clarity, one subset (denoted as ``hard-to-learn'') includes samples with top-50\% uncertainty, while the remaining samples are in the other subset (denoted as ``easy-to-learn''). We train the student model with vanilla $\mathcal{L}_{\text{KL}}$ on different training sets, and illustrate the results in Figure~\ref{fig:analysis_1}. 

Obviously, training on the ``hard-to-learn'' tokens achieves much better performance than on the ``easy-to-learn'' tokens, and even outperforms the full-data training. This indicates that \textit{tokens with more uncertainty contain more ``dark knowledge'' and are more important for KD}. Conversely, due to the shallow patterns of easy-to-learn tokens, forcing the student to learn from them might suffer from over-fitting, leading to poorer performance. More interestingly, \textit{this phenomenon seems to be more significant in larger teachers}.

\paragraph{\ding{183} \textsc{DKD} contributes more (than \textsc{TKD}) but is greatly suppressed, especially for the larger teachers.} Here, we delve into the individual effect of \textsc{TKD} and \textsc{DKD} by comparing the performance of (1) ``\textsc{TKD}-only'', (2) ``\textsc{DKD}-only'' and (3) ``\textsc{TKD}+\textsc{DKD}'' (where both are decoupled and simply added, \textit{i.e.}, ignoring the effect of \textsc{UnC}). The contrastive results among different training sets (as mentioned in \ding{182}) are listed in Table~\ref{tab:analysis_2}. 
\begin{table}[h]
\centering
\setlength{\tabcolsep}{10.8pt}
\scalebox{0.78}{
\begin{tabular}{lcccc}
\toprule
\bf{Method}   & \bf{350M} & \bf{1.3B} & \bf{2.7B} & \bf{6.7B} \\
\midrule \midrule
\multicolumn{5}{l}{\textit{1) Full data are used.}} \\
\midrule
\bf{TKD-only} & 49.19    & 48.01    & 47.21    & 48.29    \\
\bf{DKD-only} & \bf54.00    & \bf57.78    & \bf59.43    & \bf60.42    \\
\rowcolor{gray!20} \bf{TKD+DKD}  & 52.97    & 57.01    & 58.66    & 58.70   \\
\midrule
\multicolumn{5}{l}{\textit{2) Easy-to-learn tokens are used.}} \\
\midrule
\bf{TKD-only} & 39.21	&43.82	&42.37	&41.43    \\
\bf{DKD-only} & \bf48.68		&\bf54.43	&\bf58.26	&\bf60.02    \\
\rowcolor{gray!20} \bf{TKD+DKD}  & 45.59	 &44.97	&45.09	&44.66   \\
\midrule
\multicolumn{5}{l}{\textit{3) Hard-to-learn tokens are used.}} \\
\midrule
\bf{TKD-only} & 47.40	&45.15	&44.63	&48.32    \\
\bf{DKD-only} & 51.42	&58.51	&55.47	&59.88    \\
\rowcolor{gray!20} \bf{TKD+DKD}  & \bf53.26		&\bf60.49	&\bf60.60	&\bf61.47   \\
\bottomrule
\end{tabular}
}
\caption{\textbf{Comparisons of different teaching objectives.} The best results within the same training set are in \textbf{bold}.}
\label{tab:analysis_2}
\end{table}

\begin{figure}[t]
\centering
    \includegraphics[width=0.45\textwidth]{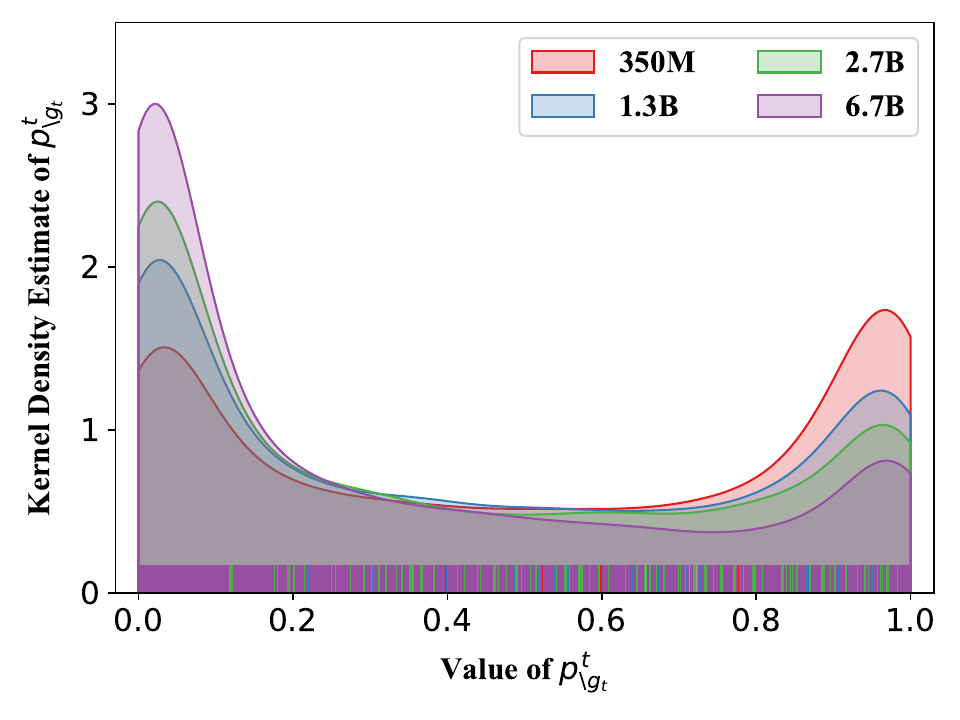}
    \caption{\textbf{Illustration of distributions of \textsc{UnC}} ($p^t_{\backslash g_t}$) among different OPT-based teachers on 100 training samples (about 10K tokens). In particular, we use the kernel density estimate for visualizing, where the larger density refers to more tokens.}
    \label{fig:analysis_2}
\end{figure}

As seen,``\textsc{DKD}-only'' outperforms the ``\textsc{TKD}-only'' among all model sizes and training sets by a large margin, indicating that the diversity-oriented knowledge is of vital importance to autoregressive KD. However, in Eq.~\ref{kd_refor}, we can find that the effect of \textsc{DKD} is suppressed by the \textsc{UnC} (ranging from 0 to 1), which might lead to the sub-optimal performance. To verify it, we further analyze the distributions of \textsc{UnC} across different model sizes. In practice, we randomly sample 100 instances from the training dataset and illustrate the distributions of \textsc{UnC} in Figure~\ref{fig:analysis_2}. It can be seen that \textsc{UnC} is generally smaller (tends to be 0) in large models than in small models, \textit{i.e.}, the larger models, the more suppressed the effect of \textsc{DKD}. This is also indicated by the results of ``\textsc{TKD}+\textsc{DKD}'', as removing the \textsc{UnC} seems to alleviate the performance degradation problem in the large models (except training on easy-to-learn tokens, where the further analyses are shown in \ding{184}). In general, these analyses prove that \textit{DKD is more important but is greatly suppressed by the \textsc{UnC} in the larger models}, which could be the main reason why a larger teacher leads to a poorer student.

\begin{figure}[t]
\centering
    \includegraphics[width=0.46\textwidth]{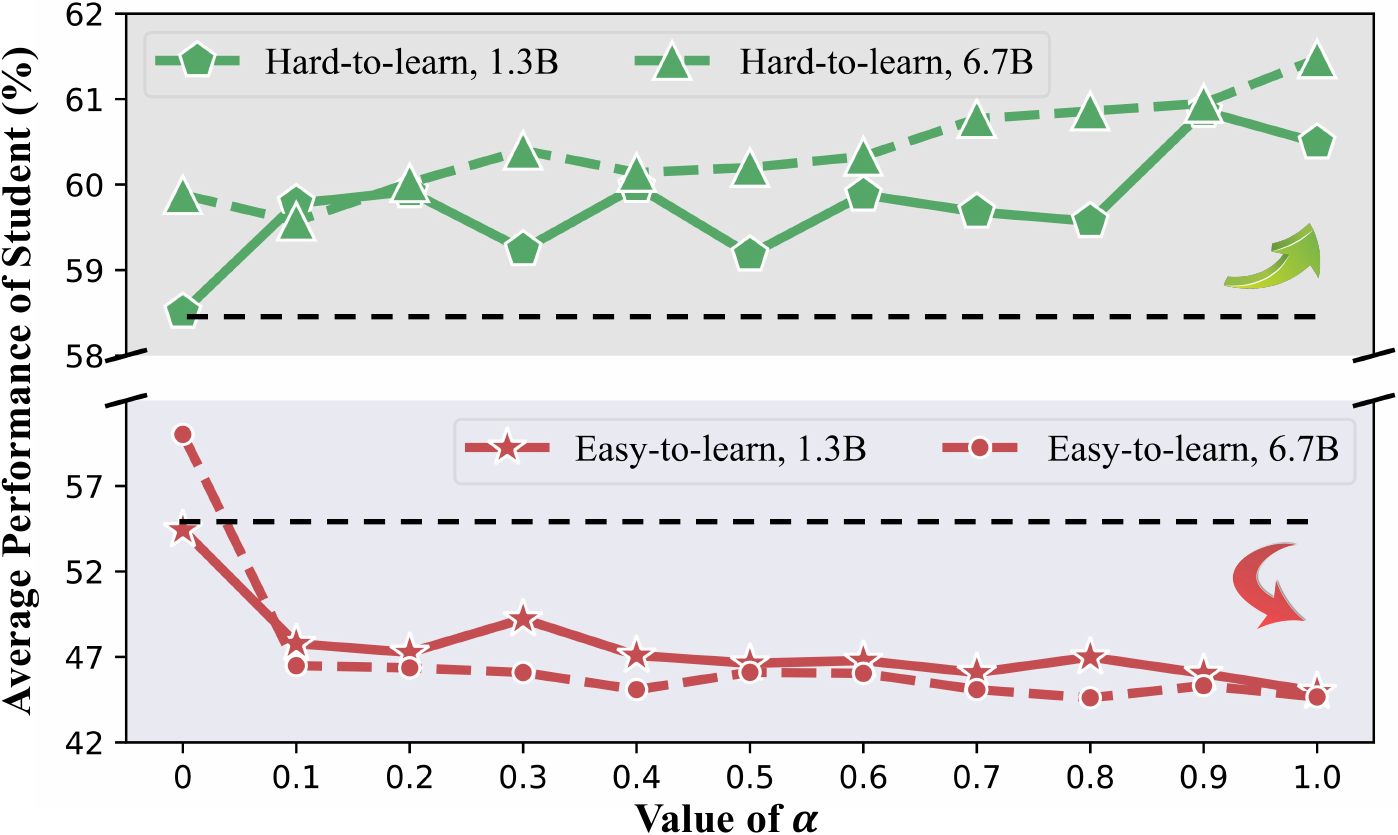}
    \caption{\textbf{Effect of TKD in different training tokens.} Here, we report the performance of students distilled with ``$\alpha\times$TKD+DKD'', where $\alpha$ is varied from 0 to 1. For ease of illustration, we only illustrate the results of using OPT-1.3B and OPT-6.7B as teachers.}
    \label{fig:analysis_3}
\end{figure}

\paragraph{\ding{184} \textsc{TKD} plays different roles in tokens with different learning difficulties.} 
We can observe an interesting phenomenon in Table~\ref{tab:analysis_2}, where adding  \textsc{TKD} upon \textsc{DKD} (``TKD+DKD'') seems to dramatically result in performance degrades when training on the easy-to-learn set, compared to the singly \textsc{DKD} (\textit{e.g.}, decreasing from 60.02\% to 44.66\%). Conversely, in the case of hard-to-learn tokens, adding \textsc{TKD} brings remarkable performance gains. These results motivate us to investigate the special effect of \textsc{TKD} on different tokens, by comparing the performance of different combinations of \textsc{TKD} and \textsc{DKD} in the setting of ``$\alpha\times$TKD+DKD''. The contrastive performance of varied $\alpha$ is illustrated in Figure~\ref{fig:analysis_3}. 
It can be seen that \textsc{TKD} indeed behaves differently in different training sets. \textsc{TKD} hurts the knowledge transfer of easy-to-learn tokens, but is beneficial to the learning of hard-to-learn tokens. We attribute it to the different learning difficulties of tokens, as \textit{the target-oriented learning on easy-to-learn tokens might damage the diversity of students}~\cite{tan2008innovation}. On the other hand, \textit{adding target-related supervision signals could reduce the learning difficulties on the hard-to-learn tokens}, thus leading to better performance.

\section{Improving Knowledge Distillation with Adaptive Teaching Modes}
\label{sec:method}
Based on the observations in \S\ref{sec:preliminaries}, we recognize that \textit{\textbf{different tokens have different teaching modes}}, and the side effect (\textit{i.e.}, problem degrades in larger teachers) of KD mainly comes from the neglect of this principle. To this end, we propose to improve the autoregressive KD with adaptive teaching modes (ATKD). In this section, we introduce the ATKD approach in detail. 

\paragraph{Motivation and Overview of ATKD.} 
In addition to the empirical findings in \S\ref{sec:preliminaries}, our ATKD is also inspired by a famous education initiative~\cite{tan2008innovation}, ``Teach Less, Learn More'', which highlights that \textit{reducing rote learning and making education more diverse and flexible can improve the quality of teaching and enhance student learning}. Intuitively, due to the large capability gap between teacher and student models, target-oriented learning of easy-to-learn tokens may encourage the student to simply mimic the teacher's shallow style but not to learn its dark knowledge~\cite{gudibande2023false}. That is, the student might fall short in generalizing to more tasks, leading to sub-optimal performance. Motivated by this, our ATKD aims to encourage the students to learn from different perspectives for different tokens. In short, ATKD skips the target-oriented teaching for the easy-to-learn tokens, and pays more attention to the learning of diverse knowledge in the hard-to-learn tokens. By doing so, our ATKD forces the student to learn more flexible and diverse knowledge, and thus improve overall performance. 

To achieve this goal, we should first obtain the easy-/hard-to-learn tokens. As mentioned in \ding{182} of \S\ref{sec:analysis}, \textsc{UnC} can effectively measure the learning difficulties of tokens, and we thus use it as a metric to select the easy-/hard-to-learn tokens. Specifically, for each mini-batch, we rank the training tokens according to \textsc{UnC} and select the top-$k$\footnote{$k$ ranges from 0\% to 100\%, and is set as 50\% by default. The analysis of $k$ can be found in \S\ref{sec:ablation}.} tokens as hard-to-learn tokens, while the others are easy-to-learn. Then, ATKD performs the KD processes with adaptive teaching modes as follows:

\paragraph{Adaptive Teaching Modes of ATKD.} 
As aforementioned, TKD and DKD contribute differently in easy-/hard-to-learn tokens. Thus, instead of using a unified teaching mode for all tokens, we use adaptive teaching modes for easy-to-learn and hard-to-learn tokens, respectively. Specifically, we decouple the TKD and DKD (\textit{i.e.}, DKD will not be suppressed by the \textsc{UnC}) to enhance the diverse learning of students. Moreover, for the easy-to-learn tokens, considering that the student can easily learn the target-class information, we skip the target-oriented teaching, \textit{i.e.}, removing TKD. On the other hand, both TKD and DKD are used for hard-to-learn tokens, as we empirically found that target-oriented teaching is essential to the learning of hard-to-learn tokens. The learning objectives of different tokens can be formulated as:
\begin{equation*}
    \begin{split}
    \mathcal{L}^{e}_{\text{KL}} &=- \sum_{t\in\mathcal{D}_e} {\text{KL}}(\mathbf{\hat{p}^t}||\mathbf{\hat{q}^t}),\\
    \mathcal{L}^{h}_{\text{KL}} &=- \sum_{t\in\mathcal{D}_h} {\text{KL}}(\mathbf{p}^t_{\mathbf b}||\mathbf{q}^t_{\mathbf b})+{\text{KL}}(\mathbf{\hat{p}^t}||\mathbf{\hat{q}^t}),
    \end{split}
    \label{kd_a}
\end{equation*}
where $\mathcal{D}_e$ and $\mathcal{D}_h$ denote the sets of easy-to-learn and hard-to-learn tokens, respectively.

Additionally, since the hard-to-learn tokens contain more informative knowledge and are more important, we adaptively combine the easy-to-learn $\mathcal{L}^{e}_{\text{KL}}$ and hard-to-learn $\mathcal{L}^{h}_{\text{KL}}$ objectives and formulate the overall learning objective of ATKD as:
\begin{equation}
    \begin{split}
    \mathcal{L}^{all}_{\text{KL}} =\lambda*\mathcal{L}^{e}_{\text{KL}}+(1-\lambda)*\mathcal{L}^{h}_{\text{KL}},
    \end{split}
    \label{kd_ours}
\end{equation}
where $\lambda$ is a weight factor to balance the different objectives, which is empirically\footnote{It should be noted that we do not finely adjust it for different datasets and tasks, but we still achieve good performance consistently. The analysis of $\lambda$ is shown in \S\ref{sec:ablation}.} set as 0.2.

\begin{table*}[t]
\setlength{\tabcolsep}{7.3pt}
\scalebox{0.8}{
\begin{tabular}{lcccccccccccc}
\toprule[1pt]
\multicolumn{1}{l}{\multirow{2}{*}{\bf Method}} & \multicolumn{3}{c}{\bf OPT-350M} & \multicolumn{3}{c}{\bf OPT-1.3B} & \multicolumn{3}{c}{\bf OPT-2.7B} & \multicolumn{3}{c}{\bf OPT-6.7B} \\ \cmidrule[1pt](lr){2-4} \cmidrule[1pt](lr){5-7} \cmidrule[1pt](lr){8-10} \cmidrule[1pt](lr){11-13} 
\multicolumn{1}{l}{} & \bf $\mathcal{S}_{\text{NLG}}$ & \bf $\mathcal{S}_{\text{NLU}}$ & \bf \underline{Avg.} & \bf $\mathcal{S}_{\text{NLG}}$ & \bf $\mathcal{S}_{\text{NLU}}$ & \bf \underline{Avg.} & \bf $\mathcal{S}_{\text{NLG}}$ & \bf $\mathcal{S}_{\text{NLU}}$ & \bf \underline{Avg.} & \bf $\mathcal{S}_{\text{NLG}}$ & \bf $\mathcal{S}_{\text{NLU}}$ & \bf \underline{Avg.} \\
\midrule[1pt]
\multicolumn{1}{l}{Teacher} & 58.33 & 20.36 & \underline{39.35} & 68.90 & 22.60 & \underline{45.75} & 74.21 & 22.28 & \underline{48.25} & 78.71 & 23.43 & \underline{51.07} \\ \midrule[0.7pt]
\multicolumn{1}{l}{Supervised KD} & 50.62 & 18.88 & \underline{34.75} & 55.57 & 17.99 & \underline{36.78} & 55.30 & 18.69 & \underline{37.00} & 55.45 & 18.33 & \underline{36.89} \\ \hdashline
\quad +\textbf{ATKD} & 52.16 & 19.58 & \underline{35.87} & 56.76 & 19.73 & \underline{38.25} & 57.26 & 19.48 & \underline{38.37} & 57.56 & 19.31 & \underline{38.43} \\
 \rowcolor{gray!20} \quad $\Delta$ ($\uparrow$) & +1.54 & +0.69 & \underline{\bf +1.12} & +1.20 & +1.74 & \underline{\bf +1.47} & +1.96 & +0.78 & \underline{\bf +1.37} & +2.11 & +0.98 & \underline{\bf +1.54} \\ \midrule[0.7pt]
\multicolumn{1}{l}{Reverse KD} & 50.54 & 18.05 & \underline{34.30} & 51.60 & 18.15 & \underline{34.87} & 51.26 & 18.56 & \underline{34.91} & 50.08 & 18.33 & \underline{34.20} \\ \hdashline
\quad +\textbf{ATKD} & 50.86 & 19.13 & \underline{34.99} & 54.40 & 19.40 & \underline{36.90} & 54.34 & 19.27 & \underline{36.80} & 54.37 & 19.16 & \underline{36.76} \\
 \rowcolor{gray!20} \quad $\Delta$ ($\uparrow$)& +0.32 & +1.08 & \underline{\bf+0.70} & +2.80 & +1.25 & \underline{\bf+2.03} & +3.08 & +0.70 & \underline{\bf+1.89} & +4.29 & +0.83 & \underline{\bf+2.56} \\ \midrule[0.7pt]
\multicolumn{1}{l}{ImitKD} & 52.27 & 18.35 & \underline{35.31} & 59.87 & 18.41 & \underline{39.14} & 59.88 & 17.46 & \underline{38.67} & 58.86 & 17.28 & \underline{38.07} \\ \hdashline
\quad +\textbf{ATKD} & 52.36 & 18.66 & \underline{35.51} & 60.76 & 19.29 & \underline{40.02} & 60.77 & 19.18 & \underline{39.97} & 62.66 & 19.56 & \underline{41.11} \\
 \rowcolor{gray!20} \quad $\Delta$ ($\uparrow$)& +0.09 & +0.31 & \underline{\bf+0.20} & +0.89 & +0.88 & \underline{\bf+0.88} & +0.89 & +1.71 & \underline{\bf+1.30} & +3.80 & +2.28 & \underline{\bf+3.04} \\ \midrule[0.7pt]
\multicolumn{1}{l}{f-distill} & 52.18 & 18.57 & \underline{35.37} & 59.74 & 19.46 & \underline{39.60} & 60.01 & 17.08 & \underline{38.55} & 59.02 & 17.80 & \underline{38.41} \\ \hdashline
\quad +\textbf{ATKD} & 52.69 & 18.80 & \underline{35.75} & 61.30 & 19.54 & \underline{40.42} & 60.70 & 19.02 & \underline{39.86} & 61.25 & 19.18 & \underline{40.22} \\
 \rowcolor{gray!20} \quad $\Delta$ ($\uparrow$)& +0.51 & +0.23 & \underline{\bf+0.37} & +1.55 & +0.08 & \underline{\bf+0.82} & +0.68 & +1.94 & \underline{\bf+1.31} & +2.23 & +1.38 & \underline{\bf+1.80} \\ \midrule[0.7pt]
\multicolumn{1}{l}{GKD} & 51.87 & 17.32 & \underline{34.59} & 61.23 & 18.77 & \underline{40.00} & 61.24 & 17.48 & \underline{39.36} & 60.59 & 16.87 & \underline{38.73} \\ \hdashline
\quad +\textbf{ATKD} & 51.90 & 18.52 & \underline{35.21} & 61.36 & 19.07 & \underline{40.21} & 62.46 & 19.21 & \underline{40.84} & 62.62 & 19.26 & \underline{40.94} \\
 \rowcolor{gray!20} \quad $\Delta$ ($\uparrow$)& +0.04 & +1.20 & \underline{\bf+0.62} & +0.13 & +0.30 & \underline{\bf+0.21} & +1.22 & +1.73 & \underline{\bf+1.48} & +2.03 & +2.39 & \underline{\bf+2.21} \\
 \bottomrule[1pt]
\end{tabular}
}
\centering
\caption{\textbf{Results (\%) of students (OPT-125M) distilling with different teachers and KD methods}. ``\underline{Avg.}'' means the average performance of $\mathcal{S}_{\text{NLG}}$ and $\mathcal{S}_{\text{NLU}}$. ``$\Delta$ ($\uparrow$)'' denotes the performance gains of ATKD against the baselines. We see that our ATKD 1) brings consistent and significant performance gains and 2) effectively alleviates the problem of performance degrades in larger teachers.}
\label{tab:main1}
\end{table*}
\section{Evaluation}
\label{sec:experiments}
\subsection{Setup}
\paragraph{Tasks and Datasets.}
We conduct extensive experiments on various LM benchmarks, covering a diversity of language generation tasks (denoted as $\mathcal{S}_{\text{NLG}}$) and language understanding tasks (denoted as $\mathcal{S}_{\text{NLU}}$). Specifically, $\mathcal{S}_{\text{NLG}}$ consists of 5 widely-used generation tasks, \textit{i.e.}, DollyEval~\cite{gu2023knowledge}, VicunaEval~\cite{chiang2023vicuna}, SelfInst~\cite{wang2022self}, Koala~\cite{geng2023koala}, and WizardLM~\cite{xu2023wizardlm} benchmarks. $\mathcal{S}_{\text{NLU}}$ includes 3 popular classification tasks, \textit{i.e.}, MMLU~\cite{hendrycks2020measuring}, Drop~\cite{dua2019drop} and BBH~\cite{suzgun2022challenging}. The details of all tasks are shown in Appendix~\ref{appendix_data}.


For evaluation on $\mathcal{S}_{\text{NLG}}$, we report the zero-shot performance by directly evaluating the instruction-following responses using the \textbf{LLM-as-judge} metric\footnote{Although some studies show that LLM-as-Judge may exhibit a certain degree of bias~\cite{gpt_bias, gpt_era}, powerful LLMs, \textit{e.g.}, ChatGPT and GPT-4, are capable of making preference determinations that are highly consistent with those of human annotators~\cite{dubois2023alpacafarm}.}. We use the same evaluation prompt in~\citet{gu2023knowledge} to instruct the ChatGPT to judge the usefulness of model responses. Notably, for each query in $\mathcal{S}_{\text{NLG}}$, we set the maximum number of output tokens as 256. As for $\mathcal{S}_{\text{NLU}}$, we follow~\citet{chen2023alpagasus} and use the code provided by~\citet{chia2023instructeval} to conduct benchmark evaluation. Specifically, we use 5-shot direct prompting and measure the exact-match score for MMLU~\cite{hendrycks2020measuring}. Regarding the Drop~\cite{dua2019drop} and BBH~\cite{suzgun2022challenging}, 3-shot direct prompting is used and exact-match scores are reported.

\paragraph{Models.} We evaluate ATKD on three types of LMs with various sizes: OPT~\cite{zhang2022opt} (\textit{student}: 125M, \textit{teachers}: 350M, 1.3B, 2.7B, 6.7B), Pythia~\cite{biderman2023pythia} (\textit{student}: 410M, \textit{teachers}: 1.4B, 2.8B), and LLaMA
(\textit{student}: 68M~\cite{miao2023specinfer}, \textit{teachers}: 1.1B~\cite{zhang2024tinyllama}, 7B~\cite{touvron2023llamav2}). Alpaca-GPT4~\cite{peng2023instruction} consisting of 52K GPT4-generated instruction-response pairs is used as training data. For teachers, we train each model with a batch size of 128 and a peak learning rate of 2e-5. For distilling students, the learning rate is selected in \{2e-4, 2e-5\} depending on model sizes, while the batch size is 256 and the maximum tokenizer length is 512. All models are trained for 3 epochs, and all experiments are conducted on 8 NVIDIA A800 (80GB) GPUs. 

\paragraph{Baselines.} We consider 5 cutting-edge KD baselines in our main experiment: Supervised KD~\cite{hinton2015distilling}, Reverse KD~\cite{gu2023knowledge}, ImitKD~\cite{lin2020autoregressive}, f-distill~\cite{wen2023f} and GKD~\cite{agarwal2023gkd}. For reference, we also report the performance of teachers as the upper bound. We use the codebase of~\citet{liu2023online} to implement these baselines and distill students.

\subsection{Compared Results}
Results of distilled models are shown in Table~\ref{tab:main1} and~\ref{tab:main2}. For ease of illustration, we only report the overall performance of $\mathcal{S}_{\text{NLG}}$ and $\mathcal{S}_{\text{NLU}}$, respectively, where the detailed results are listed in Table~\ref{tab:full_main1} and~\ref{tab:full_main2}. From these results, we can find that:

\paragraph{ATKD effectively alleviates the problem of performance degrades in larger teachers.} As seen, various baseline KD methods suffer from this problem, \textit{e.g.}, distilling OPT using GKD (\textit{1.3B: 40.00\% v.s. 6.7B: 38.73\%}). However, with the help of our ATKD, the students can generally achieve better performance in larger teachers among various baseline KD methods, \textit{i.e.}, alleviating the problem. These results can prove the effectiveness of ATKD in improving the quality of teaching.

\paragraph{ATKD brings consistent and significant performance gains among all model sizes and types.} From Table~\ref{tab:main1}, we can see that, compared with the baseline methods, our ATKD consistently achieves better performance (up to \textbf{+3.04\%} average gains) across various model sizes. Moreover, as seen in Table~\ref{tab:main2}, in addition to OPT, ATKD also works well in Pythia-family and LLaMA-family models. These results demonstrate the universality of our ATKD and indicate that ATKD has great potential to expand to more LMs.

\paragraph{ATKD is beneficial to various baseline KD methods.} In the preliminary analyses, we only conducted experiments on the typical Supervised KD. Here, we additionally investigate the combinability of ATKD and other baseline KD methods. As observed in Table~\ref{tab:main1}, ATKD can bring consistent performance gains among all baseline KD methods. For example, with the help of ATKD, Revere KD and ImitKD achieve +1.80\% and +1.36\% average performance gains, respectively.

\begin{table}[t]
\centering
\setlength{\tabcolsep}{11pt}
\scalebox{0.78}{
\begin{tabular}{lcccc}
\toprule[1pt]
\multicolumn{1}{l}{\multirow{2}{*}{\bf Method}} & \multicolumn{2}{c}{\bf Pythia-410M} & \multicolumn{2}{c}{\bf LLaMA-68M} \\ \cmidrule[1pt](lr){2-3} \cmidrule[1pt](lr){4-5}
\multicolumn{1}{l}{} & \multicolumn{1}{c}{1.4B} & \multicolumn{1}{c}{2.8B} & \multicolumn{1}{c}{1.1B} & \multicolumn{1}{c}{7B} \\ 
\midrule[1pt]
\multicolumn{1}{l}{Teacher} &67.86	&73.50  &75.23  &84.17  \\ \midrule[0.7pt]
\multicolumn{1}{l}{Supervised KD} &60.66	&59.91  &30.06  &27.94  \\ \hdashline
\quad +\textbf{ATKD} &61.81	&61.22  &31.19  &30.19  \\
  \rowcolor{gray!20} \quad $\Delta$ ($\uparrow$) &\bf +1.15	&\bf +1.31  &\bf +1.13  &\bf +2.25  \\ \midrule[0.7pt]
\multicolumn{1}{l}{Reverse KD} &55.92	&54.67  &26.15  &25.94  \\ \hdashline
\quad +\textbf{ATKD} &57.05	&57.94  &26.73  & 26.99 \\
  \rowcolor{gray!20} \quad $\Delta$ ($\uparrow$)&\bf +1.14	&\bf +3.27  &\bf +0.58  &\bf +1.05  \\ 
  \bottomrule[1pt]
\end{tabular}
}
\caption{\textbf{Results (\%) of students (Pythia-410M and LLaMA-68M)}. Due to the space limitation, we only report the results upon two typical KD baselines.}
\label{tab:main2}
\end{table}

\subsection{Ablation Study}
\label{sec:ablation}
Here, we 1) first evaluate the impact of ratio $k$, and 2) then investigate the effect of coefficient $\lambda$. Notably, we use the Supervised KD as the baseline and report the performance of OPT-125M on $\mathcal{S}_{\text{NLG}}$ tasks in this part.

\begin{figure*}[h]
\centering
    \includegraphics[width=1\textwidth]{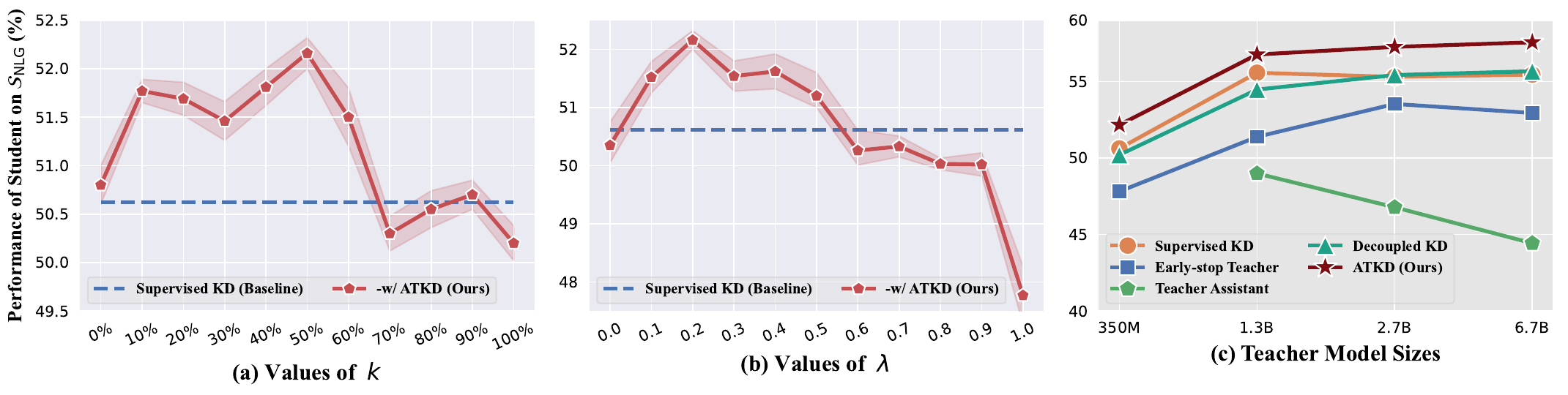}
    \caption{\textbf{(a) Effect of different ratios} (top-$k$) for selecting hard-to-learn tokens, \textbf{(b) Parameter analysis of $\alpha$} in Eq.~\ref{kd_ours}, and \textbf{(c) Comparison of different KD methods} that aim to alleviate the problem of performance degrades in larger teachers. We use the Supervised KD as the baseline and report the performance of OPT-125M on $\mathcal{S}_{\text{NLG}}$.}
    \label{fig:analysis}
\end{figure*}

\paragraph{Impact of ratio $k$.}
The ratio $k$ that is used to select the hard-to-learn tokens, is an important hyper-parameter in ATKD. In this study, we analyze its influence by evaluating the performance with different $k$ spanning from 0\% to 100\% at 10\% intervals on $\mathcal{S}_{\text{NLG}}$ tasks. Figure~\ref{fig:analysis} \textbf{(a)} illustrates the average results, in which we can find that: 1) Too large $k$ values (\textit{e.g.}, 70\%) lead to performance degradation, as many of the selected tokens are ``false'' hard-to-learn and might distort the adaptive teaching. 2)~The model's performance stably increases between 10\% and 50\%, and ATKD performs best with $k=50\%$, thus leaving as our default settings.

\begin{figure}[t]
\center
    \includegraphics[width=0.48\textwidth]{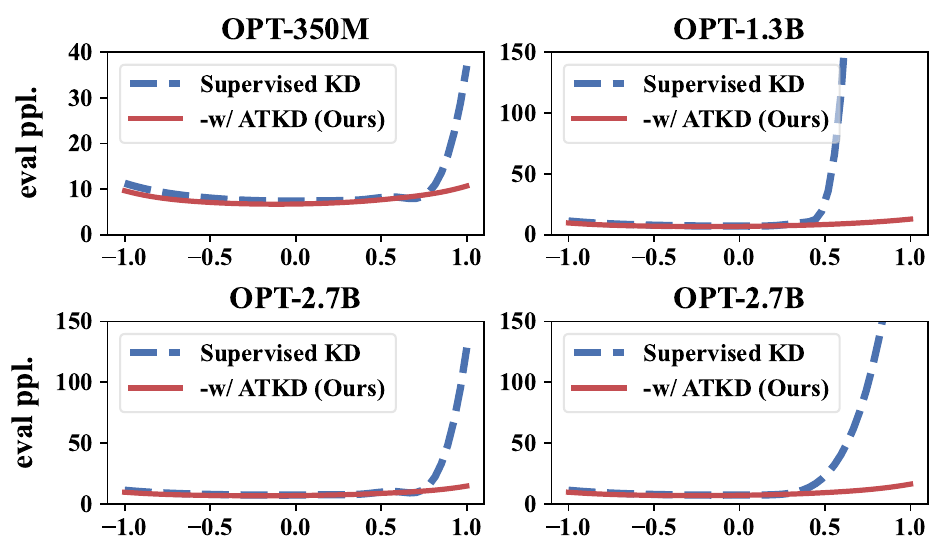}
    \caption{\textbf{1D visualization of loss landscapes of OPT-125M} distilled by different methods and teachers. The y-axis denotes the model perplexity on VicunaEval. We see that ATKD effectively smooths the loss landscape.}
    \label{fig:1d_loss}
\end{figure}

\paragraph{Impact of coefficient $\lambda$.}
The factor $\lambda$ in Eq.~\ref{kd_ours}, which is used to balance different objectives, is also needed to be investigated. Figure~\ref{fig:analysis} \textbf{(b)} illustrates the results of varied $\lambda$ ranging from 0 to 1. As seen, compared to the single learning of hard-to-learn tokens, incorporating some supervision signals from easy-to-learn tokens results in better performance. However, too large $\lambda$ values (\textit{e.g.}, 0.9) would be harmful to the effectiveness of ATKD, as paying much attention to the learning of easy-to-learn tokens might lead to overfitting. More specifically, the case of $\lambda = 0.2$ performs best, and we thereby use this setting in our experiments.

\subsection{Discussion}
\label{sec:discussion}
Here, we conduct further analyses to discuss: 1) whether ATKD outperforms the other counterparts, and 2) whether it gains better model generalization.

\paragraph{Comparison with other counterparts.} 
To the best of our knowledge, there are no existing KD methods that involve solving the problem of performance degradation for autoregressive LLMs. Thus, we compare ATKD with the related methods in the vision community: ``Early-stop Teacher''~\cite{cho2019efficacy}, ``Teacher Assistant''\footnote{We use the OPT-350M as the assistant model and only report the results distilling from teachers larger than 350M.}~\cite{mirzadeh2020improved} and ``Decoupled KD''~\cite{zhao2022decoupled}. The contrastive results are illustrated in Figure~\ref{fig:analysis} \textbf{(c)}, from which we can find that: 1) Suppressing the teacher's performance via early stopping or leveraging a smaller assistant might not be effective and even lead to worse performance, 2) Although ``Decoupled KD'' could alleviate this problem, it achieves sub-optimal performance, as it equally adopts the same teaching modes for all tokens. Takeaway: \textbf{\textit{among all methods, our ATKD can not only alleviate this problem but also bring further performance gains in a simple manner, proving its superiority.}}

\paragraph{Model Generalization.} Enforcing the student to learn more diverse knowledge could improve its generalization. To verify this conjecture, we visualize the loss landscapes of different distilled OPT-125M models on the VicunaEval task. In practice, we follow~\citet{he2021effectiveness, zhong2022improving} to plot the 1D loss curve by linear interpolation between the model weights before (denoted as $\theta_0$) and after (denoted as $\theta_1$) distilling, \textit{i.e.}, ``$\theta_1+\beta \cdot (\theta_1-\theta_0)$'', where $\beta$ is a scalar parameter that is ranged from -1 to 1. The 1D visualization results are illustrated in Figure~\ref{fig:1d_loss}, and we find that ``-w/ ATKD (Ours)'' shows a flatter and optimal property against the baseline Supervised KD. Takeaway: \textbf{\textit{These results prove that ATKD can smooth the loss landscape and improve the model generalization effectively.}}

\section{Related Works}
\label{sec:related}
 Recently, autoregressive LMs~\cite{openai2023gpt4,chowdhery2023palm,touvron2023llamav2} have shown their superior performance by solving various NLP tasks in a generative manner. Despite their success, they usually suffer from unbearable inference latency~\cite{leviathan2023fast}. To this end, several model compression approaches are proposed to reduce the model size and accelerate the inference~\cite{hinton2015distilling,jaszczur2021sparse,zhu2023zero,chen2024db}. 
 Among these efforts, KD strategy~\cite{hinton2015distilling}, which aims at training a smaller student model with the guidance of a teacher model, has attracted great attention recently~\cite{ding2021understanding,wen2023f,gu2023knowledge,agarwal2023gkd}. Although these KD methods realize promising performance when distilling (relatively) smaller LMs, they might fall short in distilling larger LMs (\textit{e.g.}, OPT-6.7B) especially when the student is of a small scale. In fact, this phenomenon has been observed in the vision community~\cite{mirzadeh2020improved,cho2019efficacy} and language understanding models~\cite{zhang2023lifting}. To alleviate this problem, a few studies including teacher assistant-based~\cite{mirzadeh2020improved} and student-friendly~\cite{cho2019efficacy,zhao2022decoupled,zhang2023lifting} distillation have been recently explored.

The above efforts are generally used for vision models or discriminative LMs, while the autoregressive KD for generative LMs is yet to be explored. To the best of our knowledge, we are the (nearly) first to alleviate the problem of performance degradation in larger autoregressive teacher LMs. Different from the previous methods that aim to directly bridge the performance gap between teacher and student, we attempt to improve the quality of teaching by exploring and addressing the limitations of existing KD objectives.

\section{Conclusion}
\label{sec:conclusion}
In this paper, we reveal and address the limitations of KD in compressing the larger autoregressive teachers. Based on a series of preliminary analyses, we find that equally adopting the same teaching modes for all tokens is sub-optimal, as learning more target-oriented knowledge of the easy-to-learn tokens might lead to overfitting and result in poor performance. To address these limitations, we improve KD with a novel adaptive teaching algorithm. It skips the target-oriented teaching for easy-to-learn tokens and pays more attention to the diverse learning of hard-to-learn tokens. Experiments show that our approach consistently and significantly improves distillation performance across all model architectures. In-depth analyses prove that our approach indeed alleviates the problem, and further improves the model generalization.

\section*{Limitations}
Our work has several potential limitations.
First, given the limited computational budget, we only validate our ATKD on up to 7B autoregressive LMs in the main experiments. Although the extra analysis in Appendx~\ref{appendix_large} shows that ATKD has the great potential to work well in distilling larger teachers, it will be more convincing if scaling up to super-large model size (\textit{e.g.}, 70B) and applying ATKD to more cutting-edge model architectures. 
On the other hand, besides the distillation performance, we believe that there are still other properties, \textit{e.g.}, training efficiency and model robustness, of LMs that can be improved by our ATKD approach, which are not fully explored in this work.

\section*{Ethics and Reproducibility Statements}
\paragraph{Ethics} 
We take ethical considerations very seriously and strictly adhere to the ACL Ethics Policy. This paper proposes an adaptive teaching algorithm to improve existing KD strategies. It aims to compress the existing larger LMs into smaller students, instead of encouraging them to learn privacy knowledge that may cause the ethical problem. Moreover, all training and evaluation datasets used in this paper are publicly available and have been widely adopted by researchers. Thus, we believe that this research will not pose ethical issues.

\paragraph{Reproducibility} In this paper, we discuss the detailed experimental setup, such as hyper-parameters and statistic descriptions. More importantly, we will publicly release our code in \url{https://github.com/WHU-ZQH/ATKD} to help reproduce the experimental results of this paper.

\section*{Acknowledgements}
We are grateful to the anonymous reviewers and the area chair for their insightful comments and suggestions.
This work was supported in part by the National Natural Science Foundation of China under Grant 623B2076, U23B2048, 62076186 and 62225113, in part by the National Key Research and Development Program of China under Grant 2023YFC2705700, and in part by the Innovative Research Group Project of Hubei Province under Grant 2024AFA017. The numerical calculations in this paper have been done on the supercomputing system in the Supercomputing Center of Wuhan University.

\bibliography{acl2024}

\begin{thebibliography}{47}
\expandafter\ifx\csname natexlab\endcsname\relax\def\natexlab#1{#1}\fi

\bibitem[{Agarwal et~al.(2024)Agarwal, Vieillard, Stanczyk, Ramos, Geist, and Bachem}]{agarwal2023gkd}
Rishabh Agarwal, Nino Vieillard, Piotr Stanczyk, Sabela Ramos, Matthieu Geist, and Olivier Bachem. 2024.
\newblock \href {https://openreview.net/forum?id=3zKtaqxLhW} {On-policy distillaiton of language models: Learning from self-generated mistakes}.
\newblock In \emph{ICLR}.

\bibitem[{Biderman et~al.(2023)Biderman, Schoelkopf, Anthony, Bradley, O’Brien, Hallahan, Khan, Purohit, Prashanth, Raff et~al.}]{biderman2023pythia}
Stella Biderman, Hailey Schoelkopf, Quentin~Gregory Anthony, Herbie Bradley, Kyle O’Brien, Eric Hallahan, Mohammad~Aflah Khan, Shivanshu Purohit, USVSN~Sai Prashanth, Edward Raff, et~al. 2023.
\newblock \href {https://proceedings.mlr.press/v202/biderman23a/biderman23a.pdf} {Pythia: A suite for analyzing large language models across training and scaling}.
\newblock In \emph{ICML}.

\bibitem[{Chen et~al.(2024)Chen, Lv, Ding, Qin, Zhou, Ding, Liu, Zhang, Guo, Liu et~al.}]{chen2024db}
Hong Chen, Chengtao Lv, Liang Ding, Haotong Qin, Xiabin Zhou, Yifu Ding, Xuebo Liu, Min Zhang, Jinyang Guo, Xianglong Liu, et~al. 2024.
\newblock \href {https://arxiv.org/abs/2402.11960} {Db-llm: Accurate dual-binarization for efficient llms}.
\newblock \emph{arXiv preprint}.

\bibitem[{Chen et~al.(2020)Chen, Wang, Utiyama, and Sumita}]{chen2020content}
Kehai Chen, Rui Wang, Masao Utiyama, and Eiichiro Sumita. 2020.
\newblock \href {https://aclanthology.org/2020.acl-main.34.pdf} {Content word aware neural machine translation}.
\newblock In \emph{ACL}.

\bibitem[{Chen et~al.(2023)Chen, Li, Yan, Wang, Gunaratna, Yadav, Tang, Srinivasan, Zhou, Huang et~al.}]{chen2023alpagasus}
Lichang Chen, Shiyang Li, Jun Yan, Hai Wang, Kalpa Gunaratna, Vikas Yadav, Zheng Tang, Vijay Srinivasan, Tianyi Zhou, Heng Huang, et~al. 2023.
\newblock \href {https://arxiv.org/pdf/2307.08701.pdf} {Alpagasus: Training a better alpaca with fewer data}.
\newblock \emph{arXiv preprint}.

\bibitem[{Chia et~al.(2023)Chia, Hong, Bing, and Poria}]{chia2023instructeval}
Yew~Ken Chia, Pengfei Hong, Lidong Bing, and Soujanya Poria. 2023.
\newblock \href {https://arxiv.org/pdf/2306.04757.pdf} {Instructeval: Towards holistic evaluation of instruction-tuned large language models}.
\newblock \emph{arXiv preprint}.

\bibitem[{Chiang et~al.(2023)Chiang, Li, Lin, Sheng, Wu, Zhang, Zheng, Zhuang, Zhuang, Gonzalez et~al.}]{chiang2023vicuna}
Wei-Lin Chiang, Zhuohan Li, Zi~Lin, Ying Sheng, Zhanghao Wu, Hao Zhang, Lianmin Zheng, Siyuan Zhuang, Yonghao Zhuang, Joseph~E Gonzalez, et~al. 2023.
\newblock \href {https://lmsys.org/blog/2023-03-30-vicuna/} {Vicuna: An open-source chatbot impressing gpt-4 with 90\%* chatgpt quality}.

\bibitem[{Cho and Hariharan(2019)}]{cho2019efficacy}
Jang~Hyun Cho and Bharath Hariharan. 2019.
\newblock \href {http://openaccess.thecvf.com/content_ICCV_2019/papers/Cho_On_the_Efficacy_of_Knowledge_Distillation_ICCV_2019_paper.pdf} {On the efficacy of knowledge distillation}.
\newblock In \emph{ICCV}.

\bibitem[{Chowdhery et~al.(2023)Chowdhery, Narang, Devlin, Bosma, Mishra, Roberts, Barham, Chung, Sutton, Gehrmann et~al.}]{chowdhery2023palm}
Aakanksha Chowdhery, Sharan Narang, Jacob Devlin, Maarten Bosma, Gaurav Mishra, Adam Roberts, Paul Barham, Hyung~Won Chung, Charles Sutton, Sebastian Gehrmann, et~al. 2023.
\newblock \href {https://www.jmlr.org/papers/volume24/22-1144/22-1144.pdf} {Palm: Scaling language modeling with pathways}.
\newblock \emph{Journal of Machine Learning Research}.

\bibitem[{Church and Hanks(1990)}]{church1990word}
Kenneth Church and Patrick Hanks. 1990.
\newblock \href {https://aclanthology.org/J90-1003.pdf} {Word association norms, mutual information, and lexicography}.
\newblock \emph{Computational linguistics}.

\bibitem[{Ding et~al.(2021)Ding, Wang, Liu, Wong, Tao, and Tu}]{ding2021understanding}
Liang Ding, Longyue Wang, Xuebo Liu, Derek~F Wong, Dacheng Tao, and Zhaopeng Tu. 2021.
\newblock \href {https://openreview.net/pdf?id=ZTFeSBIX9C} {Understanding and improving lexical choice in non-autoregressive translation}.
\newblock In \emph{ICLR}.

\bibitem[{Dua et~al.(2019)Dua, Wang, Dasigi, Stanovsky, Singh, and Gardner}]{dua2019drop}
Dheeru Dua, Yizhong Wang, Pradeep Dasigi, Gabriel Stanovsky, Sameer Singh, and Matt Gardner. 2019.
\newblock \href {https://aclanthology.org/N19-1246.pdf} {Drop: A reading comprehension benchmark requiring discrete reasoning over paragraphs}.
\newblock In \emph{NAACL-HLT}.

\bibitem[{Dubois et~al.(2023)Dubois, Li, Taori, Zhang, Gulrajani, Ba, Guestrin, Liang, and Hashimoto}]{dubois2023alpacafarm}
Yann Dubois, Xuechen Li, Rohan Taori, Tianyi Zhang, Ishaan Gulrajani, Jimmy Ba, Carlos Guestrin, Percy Liang, and Tatsunori~B Hashimoto. 2023.
\newblock \href {https://arxiv.org/pdf/2305.14387} {Alpacafarm: A simulation framework for methods that learn from human feedback}.
\newblock \emph{arXiv preprint}.

\bibitem[{Geng et~al.(2023)Geng, Gudibande, Liu, Wallace, Abbeel, Levine, and Song}]{geng2023koala}
Xinyang Geng, Arnav Gudibande, Hao Liu, Eric Wallace, Pieter Abbeel, Sergey Levine, and Dawn Song. 2023.
\newblock \href {https://bair.berkeley.edu/blog/2023/04/03/koala/} {Koala: A dialogue model for academic research}.
\newblock \emph{Blog post, April}.

\bibitem[{Gu et~al.(2023)Gu, Dong, Wei, and Huang}]{gu2023knowledge}
Yuxian Gu, Li~Dong, Furu Wei, and Minlie Huang. 2023.
\newblock \href {https://arxiv.org/pdf/2306.08543} {Knowledge distillation of large language models}.
\newblock \emph{arXiv preprint}.

\bibitem[{Gudibande et~al.(2023)Gudibande, Wallace, Snell, Geng, Liu, Abbeel, Levine, and Song}]{gudibande2023false}
Arnav Gudibande, Eric Wallace, Charlie Snell, Xinyang Geng, Hao Liu, Pieter Abbeel, Sergey Levine, and Dawn Song. 2023.
\newblock \href {https://arxiv.org/abs/2305.15717} {The false promise of imitating proprietary llms}.
\newblock \emph{arXiv preprint}.

\bibitem[{He et~al.(2021)He, Liu, Ye, Tan, Ding, Cheng, Low, Bing, and Si}]{he2021effectiveness}
Ruidan He, Linlin Liu, Hai Ye, Qingyu Tan, Bosheng Ding, Liying Cheng, Jiawei Low, Lidong Bing, and Luo Si. 2021.
\newblock \href {https://aclanthology.org/2021.acl-long.172.pdf} {On the effectiveness of adapter-based tuning for pretrained language model adaptation}.
\newblock In \emph{ACL}.

\bibitem[{Hendrycks et~al.(2020)Hendrycks, Burns, Basart, Zou, Mazeika, Song, and Steinhardt}]{hendrycks2020measuring}
Dan Hendrycks, Collin Burns, Steven Basart, Andy Zou, Mantas Mazeika, Dawn Song, and Jacob Steinhardt. 2020.
\newblock \href {https://openreview.net/pdf?id=d7KBjmI3GmQ} {Measuring massive multitask language understanding}.
\newblock In \emph{ICLR}.

\bibitem[{Hinton et~al.(2015)Hinton, Vinyals, and Dean}]{hinton2015distilling}
Geoffrey Hinton, Oriol Vinyals, and Jeff Dean. 2015.
\newblock \href {https://arxiv.org/pdf/1503.02531} {Distilling the knowledge in a neural network}.
\newblock \emph{arXiv preprint}.

\bibitem[{Jaszczur et~al.(2021)Jaszczur, Chowdhery, Mohiuddin, Kaiser, Gajewski, Michalewski, and Kanerva}]{jaszczur2021sparse}
Sebastian Jaszczur, Aakanksha Chowdhery, Afroz Mohiuddin, Lukasz Kaiser, Wojciech Gajewski, Henryk Michalewski, and Jonni Kanerva. 2021.
\newblock \href {https://proceedings.neurips.cc/paper/2021/file/51f15efdd170e6043fa02a74882f0470-Paper.pdf} {Sparse is enough in scaling transformers}.
\newblock \emph{NeurIPS}.

\bibitem[{Kim and Rush(2016)}]{kim2016sequence}
Yoon Kim and Alexander~M Rush. 2016.
\newblock \href {https://aclanthology.org/D16-1139.pdf} {Sequence-level knowledge distillation}.
\newblock In \emph{EMNLP}.

\bibitem[{Leviathan et~al.(2023)Leviathan, Kalman, and Matias}]{leviathan2023fast}
Yaniv Leviathan, Matan Kalman, and Yossi Matias. 2023.
\newblock \href {https://proceedings.mlr.press/v202/leviathan23a/leviathan23a.pdf} {Fast inference from transformers via speculative decoding}.
\newblock In \emph{ICML}.

\bibitem[{Lin et~al.(2020)Lin, Wohlwend, Chen, and Lei}]{lin2020autoregressive}
Alexander Lin, Jeremy Wohlwend, Howard Chen, and Tao Lei. 2020.
\newblock \href {https://aclanthology.org/2020.emnlp-main.494.pdf} {Autoregressive knowledge distillation through imitation learning}.
\newblock In \emph{EMNLP}.

\bibitem[{Liu et~al.(2023)Liu, Hu, Bailis, Stoica, Deng, Cheung, and Zhang}]{liu2023online}
Xiaoxuan Liu, Lanxiang Hu, Peter Bailis, Ion Stoica, Zhijie Deng, Alvin Cheung, and Hao Zhang. 2023.
\newblock \href {https://arxiv.org/pdf/2310.07177} {Online speculative decoding}.
\newblock \emph{arXiv preprint}.

\bibitem[{Lu et~al.(2023)Lu, Qiu, Ding, Zhang, Kocmi, and Tao}]{Lu2023EAPrompt}
Qingyu Lu, Baopu Qiu, Liang Ding, Kanjian Zhang, Tom Kocmi, and Dacheng Tao. 2023.
\newblock \href {https://arxiv.org/abs/2303.13809} {Error analysis prompting enables human-like translation evaluation in large language models: A case study on chatgpt}.
\newblock \emph{arXiv preprint}.

\bibitem[{Miao et~al.(2023)Miao, Oliaro, Zhang, Cheng, Wang, Wong, Chen, Arfeen, Abhyankar, and Jia}]{miao2023specinfer}
Xupeng Miao, Gabriele Oliaro, Zhihao Zhang, Xinhao Cheng, Zeyu Wang, Rae Ying~Yee Wong, Zhuoming Chen, Daiyaan Arfeen, Reyna Abhyankar, and Zhihao Jia. 2023.
\newblock \href {https://arxiv.org/pdf/2305.09781} {Specinfer: Accelerating generative llm serving with speculative inference and token tree verification}.
\newblock \emph{arXiv preprint}.

\bibitem[{Mirzadeh et~al.(2020)Mirzadeh, Farajtabar, Li, Levine, Matsukawa, and Ghasemzadeh}]{mirzadeh2020improved}
Seyed~Iman Mirzadeh, Mehrdad Farajtabar, Ang Li, Nir Levine, Akihiro Matsukawa, and Hassan Ghasemzadeh. 2020.
\newblock \href {https://ojs.aaai.org/index.php/AAAI/article/download/5963/5819} {Improved knowledge distillation via teacher assistant}.
\newblock In \emph{AAAI}.

\bibitem[{OpenAI(2023)}]{openai2023gpt4}
OpenAI. 2023.
\newblock \href {http://arxiv.org/abs/2303.08774} {Gpt-4 technical report}.

\bibitem[{Peng et~al.(2023{\natexlab{a}})Peng, Li, He, Galley, and Gao}]{peng2023instruction}
Baolin Peng, Chunyuan Li, Pengcheng He, Michel Galley, and Jianfeng Gao. 2023{\natexlab{a}}.
\newblock \href {https://arxiv.org/pdf/2304.03277.pdf?trk=public_post_comment-text} {Instruction tuning with gpt-4}.
\newblock \emph{arXiv preprint}.

\bibitem[{Peng et~al.(2023{\natexlab{b}})Peng, Ding, Zhong, Shen, Liu, Zhang, Ouyang, and Tao}]{Peng2023ChatGPT4MT}
Keqin Peng, Liang Ding, Qihuang Zhong, Li~Shen, Xuebo Liu, Min Zhang, Yuanxin Ouyang, and Dacheng Tao. 2023{\natexlab{b}}.
\newblock \href {https://aclanthology.org/2023.findings-emnlp.373} {Towards making the most of chatgpt for machine translation}.
\newblock In \emph{Findings of EMNLP}.

\bibitem[{Schwartz et~al.(2020)Schwartz, Dodge, Smith, and Etzioni}]{schwartz2020green}
Roy Schwartz, Jesse Dodge, Noah~A Smith, and Oren Etzioni. 2020.
\newblock \href {https://dl.acm.org/doi/pdf/10.1145/3381831} {Green ai}.
\newblock \emph{Communications of the ACM}.

\bibitem[{Sottana et~al.(2023)Sottana, Liang, Zou, and Yuan}]{gpt_era}
Andrea Sottana, Bin Liang, Kai Zou, and Zheng Yuan. 2023.
\newblock \href {https://doi.org/10.48550/arXiv.2310.13800} {Evaluation metrics in the era of {GPT-4:} reliably evaluating large language models on sequence to sequence tasks}.
\newblock \emph{arXiv preprint}.

\bibitem[{Srivastava et~al.(2023)Srivastava, Rastogi, Rao, Shoeb, Abid, Fisch, Brown, Santoro, Gupta, Garriga-Alonso et~al.}]{srivastava2023beyond}
Aarohi Srivastava, Abhinav Rastogi, Abhishek Rao, Abu Awal~Md Shoeb, Abubakar Abid, Adam Fisch, Adam~R Brown, Adam Santoro, Aditya Gupta, Adri{\`a} Garriga-Alonso, et~al. 2023.
\newblock \href {https://openreview.net/pdf?id=uyTL5Bvosj} {Beyond the imitation game: Quantifying and extrapolating the capabilities of language models}.
\newblock \emph{Transactions on Machine Learning Research}.

\bibitem[{Suzgun et~al.(2022)Suzgun, Scales, Sch{\"a}rli, Gehrmann, Tay, Chung, Chowdhery, Le, Chi, Zhou et~al.}]{suzgun2022challenging}
Mirac Suzgun, Nathan Scales, Nathanael Sch{\"a}rli, Sebastian Gehrmann, Yi~Tay, Hyung~Won Chung, Aakanksha Chowdhery, Quoc~V Le, Ed~H Chi, Denny Zhou, et~al. 2022.
\newblock \href {https://arxiv.org/pdf/2210.09261} {Challenging big-bench tasks and whether chain-of-thought can solve them}.
\newblock \emph{arXiv preprint}.

\bibitem[{Tan et~al.(2008)Tan, Tan, and Chua}]{tan2008innovation}
Kelvin~HK Tan, Charlene Tan, and Jude~SM Chua. 2008.
\newblock \href {https://www.researchgate.net/profile/Charlene-Tan-12/publication/281740181_Innovation_in_Education_The_'Teach_Less_Learn_More'_Initiative_in_Singapore_Schools/links/55f6625808ae6a34f663369d/Innovation-in-Education-The-Teach-Less-Learn-More-Initiative-in-Singapore-Schools.pdf} {Innovation in education: The" teach less, learn more" initiative in singapore schools}.
\newblock \emph{Innovation in education}.

\bibitem[{Touvron et~al.(2023)Touvron, Martin, Stone, Albert, Almahairi, Babaei, Bashlykov, Batra, Bhargava, Bhosale et~al.}]{touvron2023llamav2}
Hugo Touvron, Louis Martin, Kevin Stone, Peter Albert, Amjad Almahairi, Yasmine Babaei, Nikolay Bashlykov, Soumya Batra, Prajjwal Bhargava, Shruti Bhosale, et~al. 2023.
\newblock \href {https://arxiv.org/pdf/2307.09288.pdf%C3%82%C2%A0} {Llama 2: Open foundation and fine-tuned chat models}.
\newblock \emph{arXiv preprint}.

\bibitem[{Wang et~al.(2022)Wang, Kordi, Mishra, Liu, Smith, Khashabi, and Hajishirzi}]{wang2022self}
Yizhong Wang, Yeganeh Kordi, Swaroop Mishra, Alisa Liu, Noah~A Smith, Daniel Khashabi, and Hannaneh Hajishirzi. 2022.
\newblock \href {https://arxiv.org/pdf/2212.10560} {Self-instruct: Aligning language model with self generated instructions}.
\newblock \emph{arXiv preprint}.

\bibitem[{Wen et~al.(2023)Wen, Li, Du, and Mou}]{wen2023f}
Yuqiao Wen, Zichao Li, Wenyu Du, and Lili Mou. 2023.
\newblock \href {https://aclanthology.org/2023.acl-long.605.pdf} {f-divergence minimization for sequence-level knowledge distillation}.
\newblock In \emph{ACL}.

\bibitem[{Xu et~al.(2023)Xu, Sun, Zheng, Geng, Zhao, Feng, Tao, and Jiang}]{xu2023wizardlm}
Can Xu, Qingfeng Sun, Kai Zheng, Xiubo Geng, Pu~Zhao, Jiazhan Feng, Chongyang Tao, and Daxin Jiang. 2023.
\newblock \href {https://arxiv.org/pdf/2304.12244.pdf?trk=public_post_comment-text} {Wizardlm: Empowering large language models to follow complex instructions}.
\newblock \emph{arXiv preprint}.

\bibitem[{Zhang et~al.(2023)Zhang, Yang, Liu, Wang, Xian, Wang, and Song}]{zhang2023lifting}
Chen Zhang, Yang Yang, Jiahao Liu, Jingang Wang, Yunsen Xian, Benyou Wang, and Dawei Song. 2023.
\newblock \href {https://aclanthology.org/2023.acl-long.249.pdf} {Lifting the curse of capacity gap in distilling language models}.
\newblock In \emph{ACL}.

\bibitem[{Zhang et~al.(2024)Zhang, Zeng, Wang, and Lu}]{zhang2024tinyllama}
Peiyuan Zhang, Guangtao Zeng, Tianduo Wang, and Wei Lu. 2024.
\newblock \href {https://arxiv.org/pdf/2401.02385} {Tinyllama: An open-source small language model}.
\newblock \emph{arXiv preprint}.

\bibitem[{Zhang et~al.(2022)Zhang, Roller, Goyal, Artetxe, Chen, Chen, Dewan, Diab, Li, Lin et~al.}]{zhang2022opt}
Susan Zhang, Stephen Roller, Naman Goyal, Mikel Artetxe, Moya Chen, Shuohui Chen, Christopher Dewan, Mona Diab, Xian Li, Xi~Victoria Lin, et~al. 2022.
\newblock \href {https://arxiv.org/pdf/2205.01068.pdf} {Opt: Open pre-trained transformer language models}.
\newblock \emph{arXiv preprint}.

\bibitem[{Zhao et~al.(2022)Zhao, Cui, Song, Qiu, and Liang}]{zhao2022decoupled}
Borui Zhao, Quan Cui, Renjie Song, Yiyu Qiu, and Jiajun Liang. 2022.
\newblock \href {https://openaccess.thecvf.com/content/CVPR2022/papers/Zhao_Decoupled_Knowledge_Distillation_CVPR_2022_paper.pdf} {Decoupled knowledge distillation}.
\newblock In \emph{CVPR}.

\bibitem[{Zhao et~al.(2023)Zhao, Fang, Pan, Yin, and Pechenizkiy}]{gpt_bias}
Jiaxu Zhao, Meng Fang, Shirui Pan, Wenpeng Yin, and Mykola Pechenizkiy. 2023.
\newblock \href {https://doi.org/10.48550/arXiv.2312.06315} {{GPTBIAS:} {A} comprehensive framework for evaluating bias in large language models}.
\newblock \emph{arXiv preprint}.

\bibitem[{Zhong et~al.(2023)Zhong, Ding, Liu, Du, and Tao}]{zhong2023chat}
Qihuang Zhong, Liang Ding, Juhua Liu, Bo~Du, and Dacheng Tao. 2023.
\newblock \href {https://arxiv.org/abs/2302.10198} {Can chatgpt understand too? a comparative study on chatgpt and fine-tuned bert}.
\newblock \emph{arXiv preprint}.

\bibitem[{Zhong et~al.(2022)Zhong, Ding, Shen, Mi, Liu, Du, and Tao}]{zhong2022improving}
Qihuang Zhong, Liang Ding, Li~Shen, Peng Mi, Juhua Liu, Bo~Du, and Dacheng Tao. 2022.
\newblock \href {https://aclanthology.org/2022.findings-emnlp.300.pdf} {Improving sharpness-aware minimization with fisher mask for better generalization on language models}.
\newblock In \emph{Findings of EMNLP}.

\bibitem[{Zhu et~al.(2023)Zhu, Zhong, Shen, Ding, Liu, Du, and Tao}]{zhu2023zero}
Miaoxi Zhu, Qihuang Zhong, Li~Shen, Liang Ding, Juhua Liu, Bo~Du, and Dacheng Tao. 2023.
\newblock \href {https://aclanthology.org/2023.emnlp-main.696.pdf} {Zero-shot sharpness-aware quantization for pre-trained language models}.
\newblock In \emph{EMNLP}.

\end{thebibliography}

\appendix
\section{Appendix}
\label{sec:appendix}

\subsection{Details of Tasks and Datasets}
\label{appendix_data}
In this work, we conduct extensive experiments on several language generation and understanding tasks. Here, we introduce the descriptions of these tasks and datasets in detail. Firstly, we present the statistics of all evaluated datasets in Table~\ref{tab:dataset}. Then, each task is described as:

\textbf{DollyEval.} DollyEval~\cite{gu2023knowledge} is a 500-sample test set that is splitted from the databricks-dolly-15k\footnote{\url{https://github.com/databrickslabs/dolly/tree/master}} dataset.

\textbf{VicunaEval.} VicunaEval~\cite{chiang2023vicuna} contains 80 challenging questions used in the Vicuna evaluation.

\textbf{SelfInst.} SelfInst~\cite{wang2022self} is a user-oriented instruction-following test set with 252 samples.

\textbf{Koala.} This test set consists of 180 queries that~\citet{geng2023koala} source from publicly available user-written language model prompts. 

\textbf{WizardLM.} WizardLM~\cite{xu2023wizardlm} consists of 218 instances, each of which is an instruction for a specific skill, such as Math, Reasoning, Complex Formats, and so on.

\textbf{MMLU.} Massive Multitask Language Understanding (MMLU)~\cite{hendrycks2020measuring} is a popular benchmark designed to measure the multitask accuracy of LLMs, covering 57 tasks.

\textbf{Drop.} Discrete Reasoning Over Paragraphs (DROP)~\cite{dua2019drop} is a math-based reading comprehension task that requires a system to perform discrete reasoning over passages extracted from Wikipedia articles.

\textbf{BBH.} BIG-Bench Hard (BBH)~\cite{suzgun2022challenging} is a subset of 23 challenging tasks from the BIG-Bench benchmark~\cite{srivastava2023beyond}, which focuses on tasks believed to be beyond the capabilities of current language models.

\begin{table}[t]
\center
\scalebox{0.9}{
\begin{tabular}{cllr}
\toprule[1pt]
Test set                & \bf Task       & \bf \# Types       & \multicolumn{1}{l}{\bf \# Samples} \\
\midrule[1pt]
\multirow{5}{*}{$\mathcal{S}_{\text{NLG}}$} & DollyEval  & Generation     & 500                            \\
                        & VicunaEval & Generation     & 80                             \\
                        & SelfInst   & Generation     & 242                            \\
                        & Koala      & Generation     & 180                            \\
                        & WizardLM   & Generation     & 218                            \\ \midrule[0.5pt]
\multirow{3}{*}{$\mathcal{S}_{\text{NLU}}$} & MMLU       & Classification & 14,079                         \\
                        & Drop       & Classification & 9,540                          \\
                        & BBH        & Classification & 6,511 \\
\bottomrule[1pt]                        
\end{tabular}
}
\caption{\textbf{Statistics of all test sets} used in this paper.}
\label{tab:dataset}
\end{table}

\begin{table}[t]
\scalebox{0.75}{
\begin{tabular}{llcccc}
\toprule[1pt]
\bf Evaluator                & \bf Method                    & \bf 350M                 & \bf 1.3B                 & \bf 2.7B                 & \bf 6.7B                 \\ \midrule[1pt]
\multirow{2}{*}{ChatGPT} & Supervised KD             & 46.93                & 51.92                & 53.02                & 53.78                \\
&  \quad +\textbf{ATKD} & \textbf{52.75}   & \textbf{52.99}  & \textbf{53.69}                & \textbf{54.74}                \\ \midrule[0.5pt]
\multirow{2}{*}{GPT-4}   & Supervised KD             &30.09  &32.28  &32.45  &33.28  \\
                         & \quad +\textbf{ATKD} &\textbf{32.48}  & \textbf{33.21} &\textbf{33.53}  &\textbf{34.49} \\
\bottomrule[1pt]
\end{tabular}
}
\caption{\textbf{Comparison between ChatGPT-based and GPT-4-based automatic evaluators}. Here, we report the evaluation results of students (OPT-125M) on the Koala benchmark, and we can see that ChatGPT makes similar judgments to GPT-4.}
\label{tab:chatgpt}
\end{table}

\subsection{ChatGPT \textit{v.s.} GPT-4}
\label{appendix_chatgpt}
Although the GPT-4 is more commonly used as the automatic evaluator for the ``LLM-as-Judge'' metric~\cite{chen2023alpagasus, chiang2023vicuna}, it requires a much higher cost, especially for our extensive experiments. As an alternative, we use the cheaper ChatGPT as the automatic evaluator to evaluate the model responses. Here, to verify whether ChatGPT is enough to reflect the behavior of LMs, we conduct a comparative study on ChatGPT and GPT-4.  Specifically, taking the responses of OPT-125M on Koala as an example, we use the ChatGPT and GPT-4 to measure the score, respectively. As listed in Table~\ref{tab:chatgpt}, GPT-4 seems to be more strict in evaluating the model responses, as the evaluated scores of GPT-4 are generally lower than those of ChatGPT. Nevertheless, both automatic evaluators make similar judgments, \textit{i.e.}, our ATKD performs better than baselines among all model sizes. Thus, we believe that \textit{ChatGPT is enough to reflect whether the model generates a useful response, and it is credible to use ChatGPT as the automatic evaluator in this study}.

\begin{table}[t]
\scalebox{0.8}{
\begin{tabular}{lcccc}
\toprule[1pt]
\bf Method                    & Dolly & Vicuna & SelfInst  & WizardLM                 \\ \midrule[1pt]
\multicolumn{5}{l}{\textit{Distilling from larger teacher, OPT-13B}} \\ \hdashline
Supervised KD            & 58.59                & 48.88                & 55.63                & 42.44                \\
\quad +\textbf{ATKD} & \textbf{62.02}   & \textbf{56.07}  & \textbf{60.16}                & \textbf{48.37}  \\
  \rowcolor{gray!20} \quad $\Delta$ ($\uparrow$) & +3.43   & +7.19  & +4.53               & +5.93 \\
\bottomrule[1pt]
\end{tabular}
}
\caption{\textbf{Results of student (OPT-125M) distilling from larger teacher (OPT-13B)}. Here, we report several tasks from the $\mathcal{S}_{\text{NLG}}$ set. We can find that our ATKD works well in distilling larger models.}
\label{tab:large_models}
\end{table}

\subsection{Whether Our Method Works Well in Distilling Larger Models.}
\label{appendix_large}
To verify whether our method works well in the larger model settings, we conduct additional experiments using the OPT-13B teacher model. We apply the Supervised KD and our ATKD methods to distill the OPT-13B into the OPT-125M student model. Evaluation results on several $\mathcal{S}_{\text{NLG}}$ tasks are listed in Table~\ref{tab:large_models}, where we use the LLM-as-a-Judge as the metric. As seen, when using the OPT-13B model, the Supervised KD still suffers from the problem of performance degradation, as the distilled student model performs much worse than those of smaller teacher models in Table~\ref{tab:full_main1}. Conversely, our ATKD can effectively alleviate this problem and achieve much better performance (\textit{i.e.}, up to +7.19 on the VicunaEval dataset) than the baseline Supervised KD. These results indicate that our ATKD has the great potential to expand to super-large-scale model scenarios.

\begin{table*}[ht]
\center
\scalebox{0.8}{
\begin{tabular}{lcccccccccc}
\toprule[1pt]
\multirow{2}{*}{Method} & \multicolumn{5}{c}{$\mathcal{S}_{\text{NLG}}$} & \multicolumn{3}{c}{$\mathcal{S}_{\text{NLU}}$} & \multicolumn{2}{c}{Average} \\ \cmidrule[1pt](lr){2-6} \cmidrule[1pt](lr){7-9} \cmidrule[1pt](lr){10-11}
 & DollyEval & VicunaEval & SelfInst & Koala & WizardLM & MMLU & Drop & BBH & $\mathcal{S}_{\text{NLG}}$ & $\mathcal{S}_{\text{NLU}}$ \\ \midrule[1pt] \midrule[1pt]
SFT -w/o KD & 55.05 & 38.45 & 52.52 & 45.27 & 42.35 & 21.66 & 3.8 & 27.5 & 49.75 & 17.65 \\ \midrule[1pt]
\bf Teacher-OPT-350M & 64.96 & 51.09 & 61.98 & 52.13 & 46.84 & 26.03 & 6.98 & 28.08 & 58.33 & 20.36 \\ \midrule[0.5pt]
Supervised KD & 54.90 & 45.93 & 53.86 & 46.93 & 41.98 & 24.44 & 4.85 & 27.36 & 50.62 & 18.88 \\
\rowcolor{gray!20} \quad +\textbf{ATKD} & 56.14 & 43.35 & 52.98 & 52.75 & 44.88 & 24.52 & 6.94 & 27.27 & \bf 52.16 & \bf 19.58 \\ 
Reverse KD & 55.53 & 44.30 & 50.25 & 50.14 & 42.04 & 22.54 & 4.88 & 26.73 & 50.54 & 18.05 \\
\rowcolor{gray!20} \quad +\textbf{ATKD} & 54.81 & 43.68 & 52.66 & 51.05 & 42.26 & 23.74 & 6.95 & 26.70 & \bf 50.86 & \bf 19.13 \\ 
ImitKD & 55.68 & 41.30 & 54.73 & 52.51 & 45.55 & 24.06 & 4.25 & 26.73 & 52.27 & 18.35 \\
\rowcolor{gray!20} \quad +\textbf{ATKD} & 55.10 & 43.64 & 55.37 & 52.49 & 45.84 & 25.07 & 4.39 & 26.51 & \bf 52.36 & \bf 18.66 \\ 
f-distill & 56.31 & 43.52 & 52.67 & 52.69 & 44.93 & 24.71 & 4.50 & 26.49 & 52.18 & 18.57 \\
\rowcolor{gray!20} \quad +\textbf{ATKD} & 54.86 & 42.61 & 57.22 & 53.00 & 46.15 & 24.60 & 5.04 & 26.76 & \bf 52.69 & \bf 18.80 \\ 
GKD & 53.76 & 44.41 & 53.82 & 54.62 & 45.83 & 23.93 & 1.42 & 26.61 & 51.87 & 17.32 \\
\rowcolor{gray!20} \quad +\textbf{ATKD} & 54.43 & 44.35 & 53.88 & 54.79 & 44.31 & 25.40 & 2.29 & 27.88 & \bf 51.90 & \bf 18.52 \\ \midrule[1pt]
\bf Teacher-OPT-1.3B & 72.29 & 68.86 & 74.35 & 65.02 & 58.30 & 24.78 & 14.00 & 29.01 & 68.90 & 22.60 \\ \midrule[0.5pt]
Supervised KD & 60.89 & 52.35 & 57.95 & 51.92 & 44.92 & 22.27 & 4.57 & 27.13 & 55.57 & 17.99 \\
\rowcolor{gray!20} \quad +\textbf{ATKD} & 62.35 & 51.52 & 59.59 & 52.99 & 45.86 & 25.08 & 6.43 & 27.67 & \bf 56.76 & \bf 19.73 \\ 
Reverse KD & 57.16 & 46.36 & 50.75 & 50.10 & 42.94 & 23.02 & 4.22 & 27.21 & 51.60 & 18.15 \\
\rowcolor{gray!20} \quad +\textbf{ATKD} & 59.08 & 48.41 & 57.17 & 52.04 & 44.71 & 26.06 & 5.44 & 26.71 & \bf 54.40 & \bf 19.40 \\ 
ImitKD & 64.55 & 50.74 & 61.99 & 59.15 & 50.73 & 23.45 & 4.31 & 27.47 & 59.87 & 18.41 \\
\rowcolor{gray!20} \quad +\textbf{ATKD} & 65.27 & 53.70 & 63.41 & 60.00 & 50.70 & 25.76 & 4.90 & 27.20 & \bf 60.76 & \bf 19.29 \\ 
f-distill & 64.80 & 51.45 & 61.57 & 59.00 & 49.78 & 26.59 & 4.71 & 27.08 & 59.74 & 19.46 \\
\rowcolor{gray!20} \quad +\textbf{ATKD} & 65.72 & 51.56 & 62.96 & 60.72 & 53.35 & 26.58 & 4.84 & 27.21 & \bf 61.30 & \bf 19.54 \\ 
GKD & 63.48 & 56.08 & 64.73 & 61.54 & 53.83 & 25.99 & 4.42 & 25.89 & 61.23 & 18.77 \\
\rowcolor{gray!20} \quad +\textbf{ATKD} & 64.84 & 56.75 & 64.43 & 60.66 & 52.25 & 25.69 & 4.69 & 26.82 & \bf 61.36 & \bf 19.07 \\ \midrule[1pt]
\bf Teacher-OPT-2.7B & 75.64 & 74.43 & 80.99 & 74.12 & 63.39 & 24.74 & 12.86 & 29.25 & 74.21 & 22.28 \\ \midrule[0.5pt]
Supervised KD & 59.16 & 52.89 & 58.31 & 53.02 & 45.88 & 22.89 & 5.63 & 27.56 & 55.30 & 18.69 \\
\rowcolor{gray!20} \quad +\textbf{ATKD} & 62.47 & 54.47 & 60.22 & 53.69 & 46.01 & 23.83 & 6.48 & 28.12 & \bf 57.26 & \bf 19.48 \\ 
Reverse KD & 56.09 & 48.58 & 49.46 & 51.07 & 43.34 & 24.08 & 4.23 & 27.38 & 51.26 & 18.56 \\
\rowcolor{gray!20} \quad +\textbf{ATKD} & 59.79 & 50.96 & 55.73 & 50.70 & 44.54 & 24.65 & 5.76 & 27.39 & \bf 54.34 & \bf 19.27 \\ 
ImitKD & 63.30 & 57.55 & 62.98 & 59.23 & 50.01 & 22.82 & 4.50 & 25.07 & 59.88 & 17.46 \\
\rowcolor{gray!20} \quad +\textbf{ATKD} & 65.04 & 57.27 & 63.11 & 59.93 & 50.37 & 25.11 & 6.17 & 26.25 & \bf 60.77 & \bf 19.18 \\ 
f-distill & 63.78 & 58.58 & 62.79 & 58.57 & 50.00 & 22.21 & 4.40 & 24.63 & 60.01 & 17.08 \\
\rowcolor{gray!20} \quad +\textbf{ATKD} & 64.45 & 57.00 & 63.03 & 59.92 & 51.49 & 24.57 & 5.33 & 27.17 & \bf 60.70 & \bf 19.02 \\ 
GKD & 64.13 & 57.42 & 64.41 & 63.59 & 50.56 & 22.78 & 3.42 & 26.24 & 61.24 & 17.48 \\
\rowcolor{gray!20} \quad +\textbf{ATKD} & 66.84 & 60.73 & 63.23 & 63.02 & 51.72 & 25.42 & 4.57 & 27.65 & \bf 62.46 & \bf 19.21 \\ \midrule[1pt]
\bf Teacher-OPT-6.7B & 81.03 & 77.38 & 84.92 & 78.65 & 67.01 & 24.67 & 15.16 & 30.45 & 78.71 & 23.43 \\ \midrule[0.5pt]
Supervised KD & 60.01 & 49.41 & 58.22 & 53.78 & 45.51 & 23.46 & 5.43 & 26.10 & 55.45 & 18.33 \\
\rowcolor{gray!20} \quad +\textbf{ATKD} & 63.08 & 53.75 & 60.05 & 54.74 & 45.84 & 24.23 & 5.95 & 27.74 & \bf 57.56 & \bf 19.31 \\ 
Reverse KD & 53.73 & 47.33 & 49.70 & 49.50 & 43.61 & 23.95 & 4.30 & 26.73 & 50.08 & 18.33 \\
\rowcolor{gray!20} \quad +\textbf{ATKD} & 59.13 & 52.24 & 57.63 & 52.21 & 42.38 & 25.62 & 4.80 & 27.05 & \bf 54.37 & \bf 19.16 \\ 
ImitKD & 62.32 & 57.64 & 63.02 & 57.08 & 48.24 & 22.59 & 4.02 & 25.23 & 58.86 & 17.28 \\
\rowcolor{gray!20} \quad +\textbf{ATKD} & 65.07 & 58.07 & 65.93 & 63.76 & 54.29 & 25.89 & 6.68 & 26.11 & \bf 62.66 & \bf 19.56 \\ 
f-distill & 63.25 & 55.97 & 62.06 & 57.23 & 48.56 & 24.25 & 4.03 & 25.12 & 59.02 & 17.80 \\
\rowcolor{gray!20} \quad +\textbf{ATKD} & 64.51 & 59.48 & 64.04 & 62.28 & 50.48 & 25.15 & 5.57 & 26.82 & \bf 61.25 & \bf 19.18 \\ 
GKD & 64.37 & 58.47 & 61.63 & 62.19 & 50.23 & 22.03 & 3.53 & 25.04 & 60.59 & 16.87 \\
\rowcolor{gray!20} \quad +\textbf{ATKD} & 66.68 & 60.87 & 65.29 & 63.19 & 50.51 & 25.84 & 4.36 & 27.58 & \bf 62.62 & \bf 19.26 \\
\bottomrule[1pt]
\end{tabular}
}
 \caption{\textbf{Full results of Table~\ref{tab:main1}, \textit{i.e.}, performance of student (OPT-125M) on $\mathcal{S}_{\text{NLG}}$ and $\mathcal{S}_{\text{NLU}}$ across different teachers and KD methods.} ``Average'' denotes the average results of $\mathcal{S}_{\text{NLG}}$ and $\mathcal{S}_{\text{NLU}}$, and ``SFT -w/o KD'' refers to the results of the vanilla student that is tuned on the ground-truth data. Better results among baseline KD methods and ours are in \textbf{bold}.}
 \label{tab:full_main1}
\end{table*}

\begin{table*}
\begin{minipage}{0.5\textwidth}
\center
\scalebox{0.8}{
\begin{tabular}{lccccc}
\toprule[1pt]
\multirow{2}{*}{Method} & \multicolumn{5}{c}{\bf $\mathcal{S}_{\text{NLG}}$, Pythia-410M}\\ \cmidrule[1pt](lr){2-6}
 & Dolly & Vicuna & SelfInst & Koala & WizardLM  \\ \midrule[1pt] \midrule[1pt]
SFT-w/o KD & 61.81 & 57.38 & 60.62 & 50.67 & 50.06  \\ \midrule[1pt]
\bf Teacher-1.4B/-1.1B & 69.51 & 73.13 & 69.59 & 65.17 & 62.44  \\ \midrule[0.5pt]
Supervised KD & 62.62 & 64.61 & 63.47 & 56.36 & 55.15  \\
\rowcolor{gray!20} \quad +\textbf{ATKD} (Ours) & 63.87 & 62.70 & 65.64 & 59.13 & 54.72  \\
Reverse KD & 58.82 & 56.17 & 58.87 & 53.17 & 48.15  \\
\rowcolor{gray!20} \quad +\textbf{ATKD} (Ours) & 61.14 & 57.80 & 57.30 & 53.85 & 49.77 \\
\midrule[1pt]
\bf Teacher-2.8B/-7B & 75.84 & 76.63 & 72.99 & 70.95 & 69.63  \\ \midrule[0.5pt]
Supervised KD & 61.10 & 60.38 & 63.51 & 56.99 & 55.43  \\
\rowcolor{gray!20} \quad +\textbf{ATKD} (Ours) & 63.37 & 64.31 & 63.06 & 59.22 & 54.79  \\
Reverse KD & 58.80 & 54.99 & 53.74 & 52.61 & 47.79  \\
\rowcolor{gray!20} \quad +\textbf{ATKD} (Ours) & 61.21 & 63.23 & 56.38 & 57.23 & 50.81  \\
\bottomrule[1pt]
\end{tabular}
}
\caption{Results of Pythia-410M.}
\label{sub_tab:pythia}
\end{minipage}\hspace{8mm}
\begin{minipage}{0.5\textwidth}
\center
\scalebox{0.8}{
\begin{tabular}{ccccc}
\toprule[1pt]
 \multicolumn{5}{c}{\bf $\mathcal{S}_{\text{NLG}}$, LLaMA-68M}  \\ \cmidrule[1pt](lr){1-5}
 Dolly & Vicuna & SelfInst & Koala & WizardLM   \\ \midrule[1pt] \midrule[1pt]
 26.37 & 26.67 & 28.37 & 27.27 & 23.72  \\ \midrule[1pt]
 78.82 & 77.50 & 75.02 & 72.01 & 69.08  \\ \midrule[0.5pt]
29.63 & 28.74 & 31.51 & 31.84 & 28.45  \\
 \rowcolor{gray!20}30.29 & 29.50 & 34.65 & 33.25 & 28.34  \\
 25.69 & 25.74 & 27.98 & 29.23 & 22.78  \\
 \rowcolor{gray!20}26.02 & 25.31 & 28.95 & 29.23 & 24.35 \\
\midrule[1pt]
86.50 & 83.25 & 83.70 & 84.18 & 79.68  \\ \midrule[0.5pt]
 27.27 & 28.31 & 29.26 & 30.02 & 26.15  \\
 \rowcolor{gray!20}30.08 & 28.95 & 30.97 & 32.09 & 28.47  \\
25.65 & 25.11 & 27.85 & 28.03 & 23.05  \\
 \rowcolor{gray!20}26.70 & 27.38 & 28.35 & 29.53 & 23.91  \\
\bottomrule[1pt]
\end{tabular}
}
\caption{Results of LLaMA-68M.}
\label{sub_tab:llama}
\end{minipage}
\caption{\textbf{Full results of Table~\ref{tab:main2}, \textit{i.e.}, performance of students (Pythia-410M, Table~\ref{sub_tab:pythia} and LLaMA-68M, Table~\ref{sub_tab:llama}) on $\mathcal{S}_{\text{NLG}}$.} Notably, for Pythia-410M, we use the Pythia-1.4B/2.8B as teachers, while LLaMA-1.1B/7B are used as teachers for LLaMA-68M.}
\label{tab:full_main2}
\end{table*}

\end{document}